\documentclass[10pt,twocolumn,letterpaper]{article}

\usepackage{iccv}
\usepackage{times}
\usepackage{epsfig}
\usepackage{graphicx}
\usepackage{amsmath}
\usepackage{amssymb}
\usepackage{multirow}
\usepackage{algorithm}
\usepackage{algorithmic}
\usepackage[yyyymmdd,hhmmss]{datetime}

\usepackage{overpic}
\usepackage{makecell}
\usepackage{boldline}

\usepackage{siunitx}

\usepackage{array}
\newcolumntype{P}[1]{>{\centering\arraybackslash}p{#1}}
\newcolumntype{M}[1]{>{\centering\arraybackslash}m{#1}}
\newcolumntype{R}[1]{>{\raggedleft\arraybackslash}m{#1}}

\usepackage{enumitem}
\setlist{nolistsep}

\def\wrt{\emph{w.r.t.}}

\def\g{{\mathbf g}}
\def\x{{\mathbf x}}
\def\z{{\mathbf z}}

\def\m{{\mathbf m}}
\def\0{{\mathbf 0}}

\def\P{{\mathcal P}}

\def\sgn{{\rm sgn}}

\def\bbox {{\mathbf{b}}}
\def\p {{\mathbf{p}}}

\def\tt{{\boldsymbol \theta}}

\def\Loss{\mathcal{L}}

\def\lc {{\rm loss}_{\rm cls}}
\def\ll {{\rm loss}_{\rm loc}}

\usepackage[dvipsnames]{xcolor}
\definecolor{lgray}{gray}{0.8}

\def\wrt{\emph{w.r.t.}}
\newcommand{\alg}[1]{\textbf{\texttt{#1}}}
\newcommand{\salg}[1]{{\small\textbf{\texttt{#1}}}}

\newcommand{\name}[1]{{\small{\textsf{#1}}}}

\usepackage{hyperref}

\usepackage{enumitem}
\usepackage{xcolor}

\iccvfinalcopy % *** Uncomment this line for the final submission

\ificcvfinal\fi

\begin{document}
%\newrefcontext[labelprefix=A]

%%%%%%%%% TITLE
\title{Towards Adversarially Robust Object Detection}

\author{Haichao Zhang  \quad  Jianyu Wang \\
	{\fontsize{11}{12} \selectfont \begin{tabular}{c}
			{Baidu Research, \,Sunnyvale USA}
		\end{tabular}} \\
		{\fontsize{9.5}{10}	\texttt{hczhang1@gmail.com\, wjyouch@gmail.com}}\\
	}

\maketitle

%%%%%%%%% ABSTRACT
\begin{abstract}
Object detection is an important vision task and has emerged as an indispensable component in many vision system, rendering its robustness as an  increasingly important performance factor for practical applications.
While object detection models have been demonstrated to be vulnerable against adversarial attacks by many recent works, very few efforts have been devoted to improving their robustness.
In this work, we take an initial attempt towards this direction. We first revisit and systematically analyze object detectors and many recently developed attacks from the perspective of model robustness. We then present a multi-task learning perspective of object detection and identify an asymmetric role of  task losses. We further develop an adversarial training approach which can leverage the multiple sources of attacks for improving the robustness of detection models.
Extensive experiments on PASCAL-VOC and MS-COCO verified the effectiveness of the proposed approach.
  
\end{abstract}

%%%%%%%%% BODY TEXT

\section{Introduction}

Deep learning models have been widely applied to many vision tasks such as classification~\cite{VGG, deeper, resnet} and object detection~\cite{rcnn, fast_rcnn, ssd, yolo, faster_RCNN, Cascade_RCNN}, leading to state-of-the-art performance.
However, one impeding factor of deep learning models is their issues with robustness.
It has been shown that deep net-based classifiers are vulnerable to  adversarial attack~\cite{szegedy2013intriguing,FGSM}, \emph{i.e.}, 
there exist adversarial examples
that are slightly modified but visually indistinguishable version of the original images that cause the classifier to generate incorrect predictions~\cite{moosavi2015deepfool,carlini2016towards}.
Many efforts have been devoted to improving the robustness of classifiers~\cite{metzen2017detecting,meng2017magnet, xie2017mitigating,guo2017countering,liao2017defense,samangouei2018defensegan,song2018pixeldefend,prakash2018deflecting,liu2017towards}.

Object detection is a computer vision technique that deals with detecting instances of semantic objects in images~\cite{face_detector, HOG, dpm}.
It is a natural generalization of the vanilla classification task as it outputs not only the object label as in classification but also the location.
Many successful object detection approaches have been developed during the past several years~\cite{rcnn, fast_rcnn, faster_RCNN, ssd, yolo} and object detectors powered by deep nets have emerged as an indispensable component in many vision systems of real-world applications.

\begin{figure}
	\centering
	\begin{overpic}[viewport=160 60 640 555, clip, height=4cm]{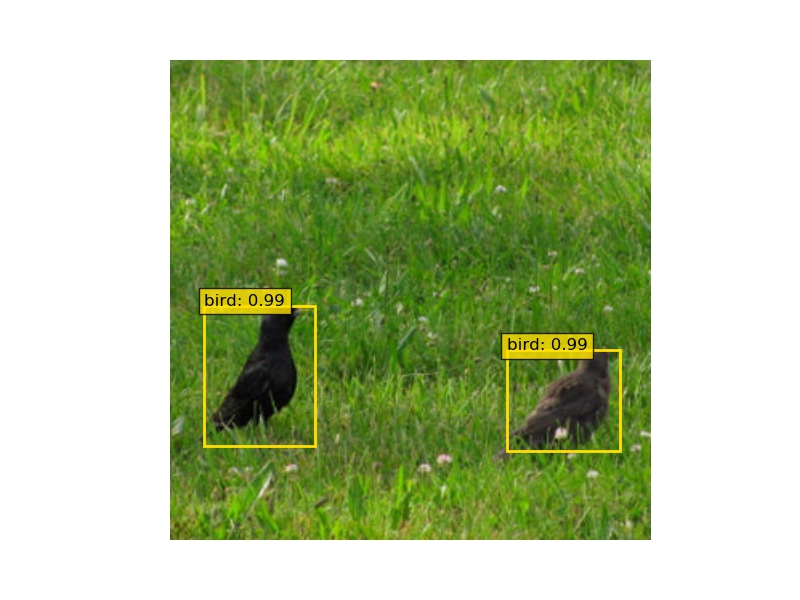}
		\put(-7,40){\sffamily\textcolor{black}{{\scalebox{0.8}{\rotatebox{90}{clean}}}}}
		\put(25,100){\sffamily\textcolor{black}{{\scalebox{0.8}{\rotatebox{0}{standard detector}}}}}
	\end{overpic}
	\begin{overpic}[viewport=160 60 640 555, clip, height=4cm]{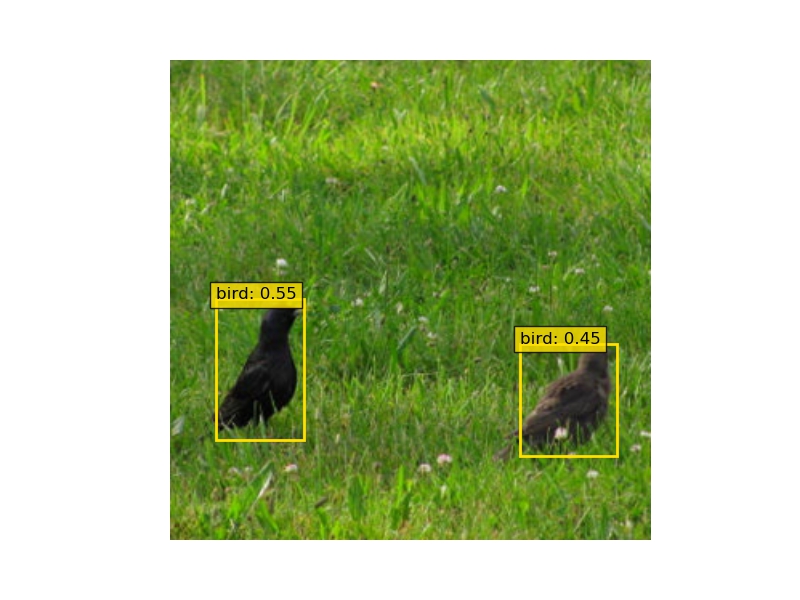}
		\put(25,100){\sffamily\textcolor{black}{{\scalebox{0.8}{\rotatebox{0}{robust detector}}}}}
	\end{overpic}
	% adv image
	\begin{overpic}[viewport=160 60 640 555, clip, height=4cm]{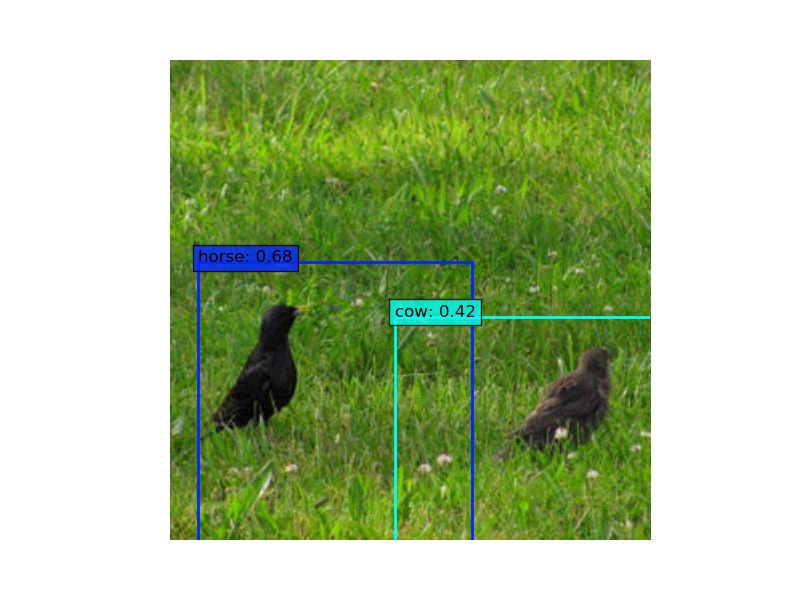}
		\put(-7,30){\sffamily\textcolor{black}{{\scalebox{0.8}{\rotatebox{90}{adversarial}}}}}
	\end{overpic}
	\begin{overpic}[viewport=160 60 640 555, clip, height=4cm]{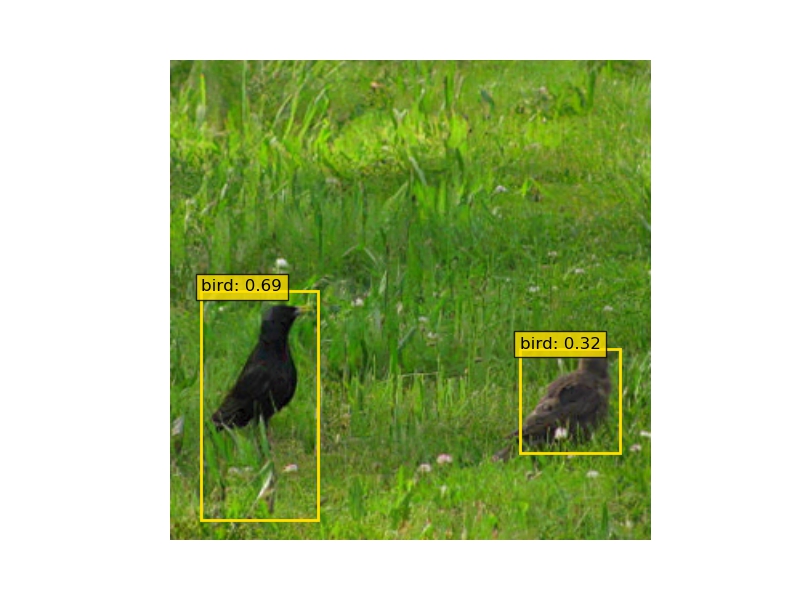}
	\end{overpic}
	\caption{\textbf{Standard v.s. robust detectors} on clean and adversarial images. The adversarial image is produced using PDG-based detector attacks~\cite{BMVC_RAP, madry2017towards} with perturbation budget 8 (out of 256). The standard model~\cite{ssd} fails completely  on the adversarial image while the robust model can produce reasonable detection results.}
	\label{fig:illustration}
	\vspace{-0.2in}
\end{figure}

Recently, it has been shown that object detectors can also be attacked by maliciously crafted inputs~\cite{DAG, detector_fool, BMVC_RAP, ShapeShifter,  trans_attack, phy_attack, dpatch, background_patch} (\cf Figure~\ref{fig:illustration}).
Given its critical role in applications such as  surveillance and autonomous driving, it is important to investigate approaches for defending object detectors against various adversarial attacks.
However, while many works have shown it is possible to attack a detector, it remains largely unclear whether it is possible to improve the robustness of the detectors and what is the practical approach for that.
This work servers as an initial attempt to bridge this gap towards this direction.
We show that it is \emph{possible} to improve the robustness of the object detector \wrt various types of attacks and propose a practical approach for achieving this, by generalizing the adversarial training framework from classification to detection.
%A comparison between the results of standard and robust detectors are shown in Figure~\ref{fig:illustration}.

The contribution of this paper is threefold:
\emph{\textbf{i}})~we provide a categorization and analysis of different attacks for object detectors, revealing their shared underlying mechanisms; 
\emph{\textbf{ii}})~we highlight and analyze the interactions between different tasks losses and their implication on robustness;
\emph{\textbf{iii}})~we generalize the adversarial training framework from classification to detection and  develop an  adversarial training approach that can properly handle the interactions between task losses for improving detection robustness.

\section{Related Work}
{\flushleft \textbf{Attacks and Adversarial Training for Classification}.} 
Adversarial examples have been investigated for general learning-based classifiers before~\cite{BiggioR18}. As a learning-based model, deep networks are also vulnerable to adversarial examples~\cite{szegedy2013intriguing,nguyen2015deep}.
Many variants of attacks~\cite{FGSM, moosavi2015deepfool,carlini2016towards} and defenses~\cite{metzen2017detecting,meng2017magnet, xie2017mitigating,guo2017countering,liao2017defense,liu2017towards,samangouei2018defensegan,song2018pixeldefend,prakash2018deflecting,athalye2018obfuscated} have been developed. 
Fast gradient sign method (FGSM)~\cite{FGSM}  and Projective Gradient Descend (PGD)~\cite{madry2017towards} are two representative approaches for white-box adversarial attack generation.
Adversarial training~\cite{FGSM, kurakin2016scale, tramer2017ensemble, madry2017towards} is one of the effective defense method against adversarial attacks.
It achieves robust model training by solving a minimax problem, where the inner maximization  generates attacks according to the current model parameters while the outer optimization  minimize the training loss \wrt the model parameters~\cite{FGSM,madry2017towards}.

{\flushleft \textbf{Object Detection and Adversarial Attacks}.} 
Many successful object detection approaches have been developed during the past several years, including one-stage~\cite{ssd, yolo} and two-stage variants~\cite{rcnn, fast_rcnn, faster_RCNN}.
Two stage detectors refine proposals from the first stage by one or multiple refinement steps~\cite{faster_RCNN, Cascade_RCNN}.
We focus on one-stage detectors in this work due to its essential role in different variants of detectors.
A number of attacks for object detectors have been developed very recently~\cite{DAG, detector_fool, ShapeShifter, phy_attack, trans_attack,BMVC_RAP,background_patch, dpatch}.
\cite{DAG} extends the attack generation method from classification to detection and demonstrates that it is possible to attack objectors using a designed classification loss.
Lu \emph{et al.} generate adversarial examples that fool detectors for  stop sign and face detections~\cite{detector_fool}.
\cite{ShapeShifter} develops physical attacks for Faster-RCNN~\cite{faster_RCNN} and adapts the expectation-over-transformation idea for generating physical attacks that remain effective under various transformations such as viewpoint variations.
\cite{BMVC_RAP} proposes to attack the region-proposal network (RPN) with a specially designed hybrid loss incorporating both classification and localization terms.
Apart from the full images, it is also possible to attack detectors by restricting the attacks to be within a local region~\cite{background_patch, dpatch}.

\begin{figure}
	\centering
	\includegraphics[width=3in]{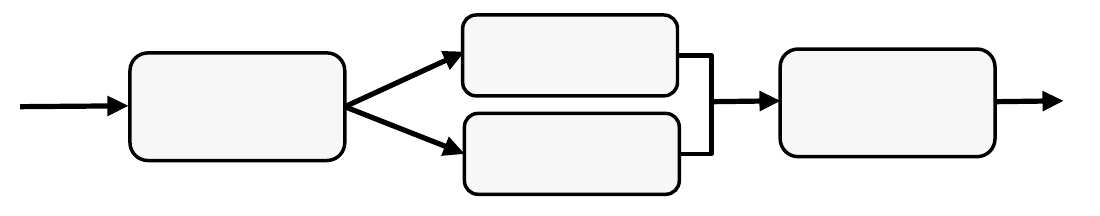}
	\put(-185,21){\small base-net}
	\put(-173,13){\small {\scriptsize $\tt_b$}}
	\put(-122,32){\scriptsize classification}
	\put(-109,25.5){\small {\scriptsize $\tt_c$}}
	\put(-120,13){\scriptsize localization}
	\put(-109,6.5){\small {\scriptsize $\tt_l$}}
	\put(-50,19){\footnotesize NMS}
	\caption{\textbf{One-stage detector architecture}. 
	A base-net (w. para. $\tt_b$) is shared by classification (w. para.  $\tt_c$) and localization (w. para. $\tt_l$) tasks.
	$\tt = [\tt_b, \tt_c, \tt_l]$ denotes the full parameters for the detector. 
	For training, the NMS module is removed and task losses are appended for classification and localization respectively.} 
	\label{fig:ssd_chart}
\end{figure}
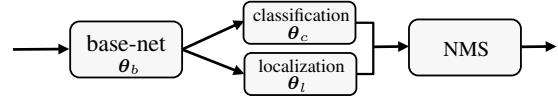

\section{Object Detection and Attacks Revisited}
We  revisit object detection and discuss the connections between many variants of attacks developed recently.

\subsection{Object Detection as Multi-Task Learning}
\label{sec:revisited}
An object detector $f(\x) \rightarrow \{\p_k, \bbox_k\}_{k=1}^K$ takes an image $\x \in [0, 255]^n$ as input and outputs a varying number of $K$ detected objects, each represented by a probability vector $\p_k\!\in\!\mathbb{R}^{C}$ over $C$ classes (including background) and  a bounding box $\bbox_k\!=\![x_k, y_k, w_k, h_k]$. Non-maximum suppression (NMS)~\cite{NMS} is applied to remove redundant detections for the final detections (\cf Figure~\ref{fig:ssd_chart}).

For training, we parametrize the detector $f(\cdot)$ by $\tt$. Then the training of the detector boils down to the estimation of $\tt$ which can be formulated as follows:
\begin{eqnarray}\label{eq:ssd}
\begin{split}
&\quad \min_{\tt} \mathbb{E}_{(\x,\{y_k, \bbox_k\}) \sim \mathcal{D}} \,\Loss(f_{\tt}(\x), \{y_k, \bbox_k\}) .
\end{split}
\end{eqnarray}
$\x$ denotes the training image and $\{y_k, \bbox_k\}$ the ground-truth (class label $y_k$ and the  bounding box $\bbox_k$) sampled from the dataset $\mathcal{D}$. 
 We will drop the expectation over data and present subsequent derivations with a single example to avoid notation clutter  without loss of generality as follows:
 \begin{eqnarray}\label{eq:ssd_single}
 	\begin{split}
 		&\quad \min_{\tt}  \Loss(f_{\tt}(\x), \{y_k, \bbox_k\}).
 	\end{split}
 \end{eqnarray}
$\Loss(\cdot)$  is a loss function measuring the difference between the output of $f_{\tt}(\cdot)$ and the ground-truth and the minimization of it (over the dataset) leads to a proper estimation of $\tt$.
In practice, it is typically instantiated as a combination of classification loss and localization loss as follows~\cite{ssd, yolo}:
\begin{eqnarray}\label{eq:ssd_task_loss}
	\begin{split}
		&\min_{\tt}\lc(f_{\tt}(\x), \{y_k, \bbox_k\}) \!+\! \ll(f_{\tt}(\x), \{y_k, \bbox_k\}).
		%&=& \min_{\tt}\Loss_{\rm cls}(f(\x), \{y_i\}_{i=1}^K) + \Loss_{\rm loc}(f(\x), \{b_i\}_{i=1}^K),
	\end{split}
\end{eqnarray}
As shown in Eqn.(\ref{eq:ssd_task_loss}),
the classification and localization tasks share some intermediate computations including the base-net (\emph{c.f.} Figure~\ref{fig:ssd_chart}).
However, they use different parts of the output from $f_{\tt}(\cdot)$ for computing losses emphasizing on different aspects, \emph{i.e.}, classification and localization performance respectively.
This is a design choice for sharing feature and computation for potentially relevant tasks~\cite{ssd, yolo},
which is essentially an instance of \emph{multi-task learning}~\cite{multitask_learning}.

\begin{table}[t!]
	\centering
	\begin{tabular}{ |c V{1.5} M{0.3cm}|M{0.3cm}|M{0.3cm}|M{0.3cm}|}
		\hline
		\multirow{3}{*}{Attacks for Object Detection}
		& \multicolumn{4}{c|}{Components} \\
		\cline{2-5} 
		& \multicolumn{2}{c|}{$\lc$} & \multicolumn{2}{c|}{$\ll$} \\
		\cline{2-5} 
		& T & N & T & N \\
		\Xhline{2\arrayrulewidth}
		\alg{ShapeShifter} \cite{ShapeShifter} & \checkmark  & & & \\
		\hline
		\alg{DFool} \cite{detector_fool}, \alg{PhyAttack} \cite{phy_attack} &   & \checkmark & & \\
		\hline 
		\alg{DAG} \cite{DAG}, \alg{Transfer} \cite{trans_attack} & \checkmark & \checkmark & & \\
		\hline
		\alg{DPatch} \cite{dpatch} &  \checkmark & &\checkmark & \\
		\hline
		\alg{RAP} \cite{BMVC_RAP} &  & \checkmark &\checkmark & \\
		\hline
		\alg{BPatch} \cite{background_patch} && \checkmark &  &\checkmark \\
		\hline
	\end{tabular}
	\caption{Analysis of existing attack methods for object detection. ``T" denotes ``targeted attack" and ``N" for ``non-targeted attack". }
	\label{tab:attack_analysis}
	%\vspace{-0.15in}
	\vspace{-0.1in}
\end{table}

\subsection{Detection Attacks Guided by Task Losses}
Many different attack methods for object detectors have been developed very recently~\cite{DAG, detector_fool, ShapeShifter, phy_attack, trans_attack,BMVC_RAP,background_patch, dpatch}.
Although there are many differences in the formulations of these attacks, when viewed from the \emph{multi-task learning} perspective as pointed out in Section~\ref{sec:revisited}, they have the same framework and design principle:
\emph{an attack to a detector can be achieved by utilizing variants of individual task losses or their combinations}.
This provides a common grounding for understanding and comparing different attacks for object detectors. 
From this perspective, we  can categorize existing attack methods as  in Table~\ref{tab:attack_analysis}.
It is clear that some methods use classification loss~\cite{ShapeShifter,  detector_fool,  phy_attack, DAG, trans_attack} while other methods also incorporated localization loss~\cite{dpatch, BMVC_RAP,background_patch}.
There are two perspectives for explaining the effectiveness of individual task loss in generating attacks:
\emph{\textbf{i}}) the classification and localization tasks share a common base-net, implying that the weakness in the base-net will be shared among all tasks built upon it;
\emph{\textbf{ii}})  while the classification and localization outputs have dedicated branches for each task beyond the shared base-net, 
they are coupled in the testing phase due to the usage of NMS, which jointly use class scores and bounding box locations for redundant prediction pruning.

Although many attacks have been developed and it is possible to come up with new combinations and configurations following the general principle, there is a lack of understanding on the \emph{role of individual components} in model robustness.
 Filling this gap one of our contributions which will naturally lead to our robust training method for object detectors as detailed in the sequel.

\section{Towards Adversarially Robust Detection}

\begin{figure}
	\centering
	\begin{overpic}[viewport=45 2 534 400, clip, width=1.6in]{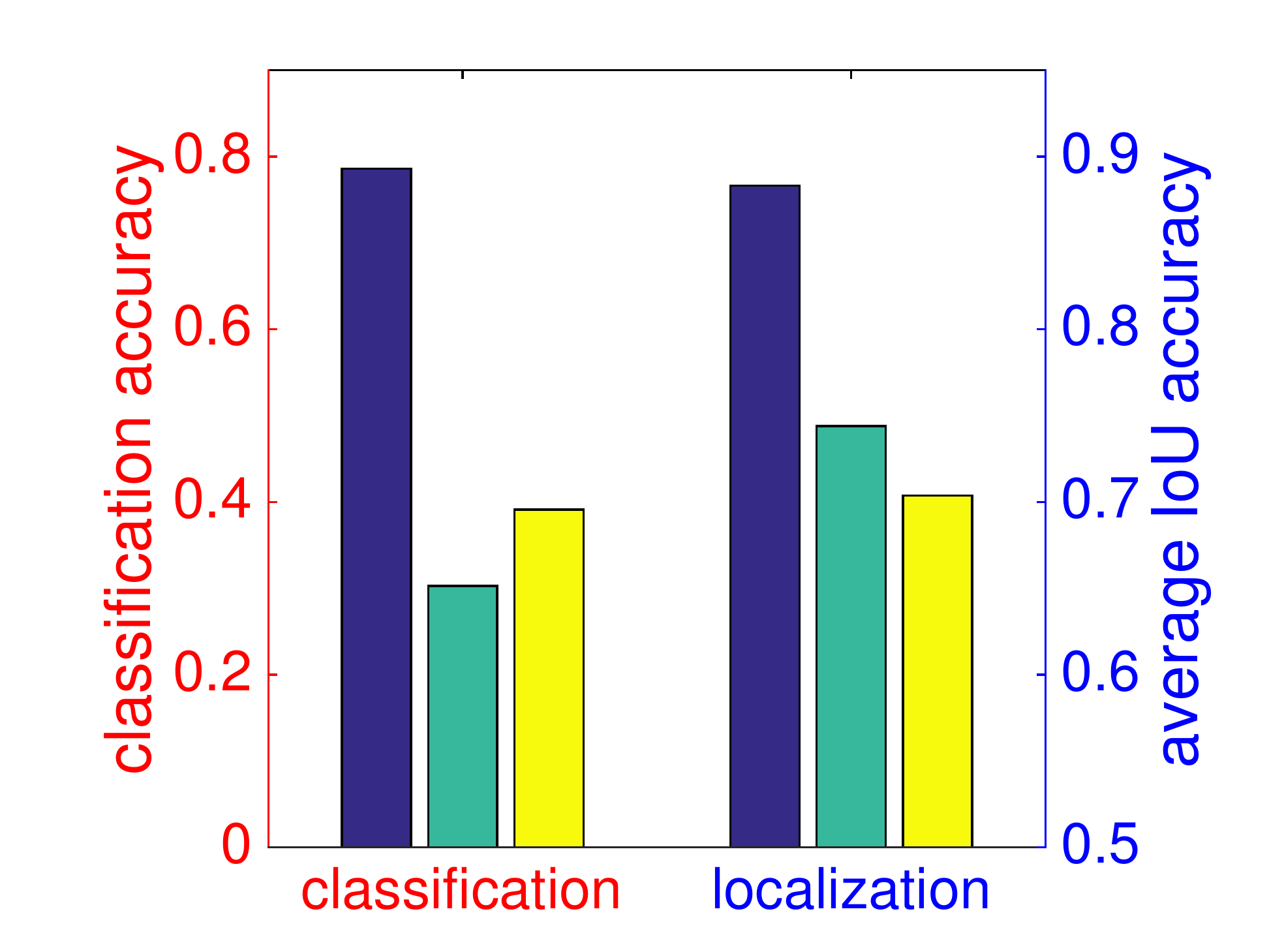}
			\put(34,72){\includegraphics[width=3.8cm]{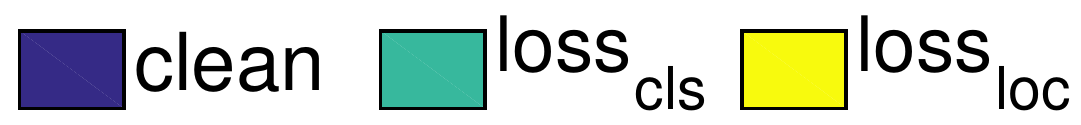}}	
	\put(1,2){\sffamily\textcolor{black}{{\scalebox{0.7}{\rotatebox{0}{(a)}}}}}
	\end{overpic}
	%\hspace{0.01in}
	\begin{overpic}[viewport=20 10 510 410, clip, width=1.6in]{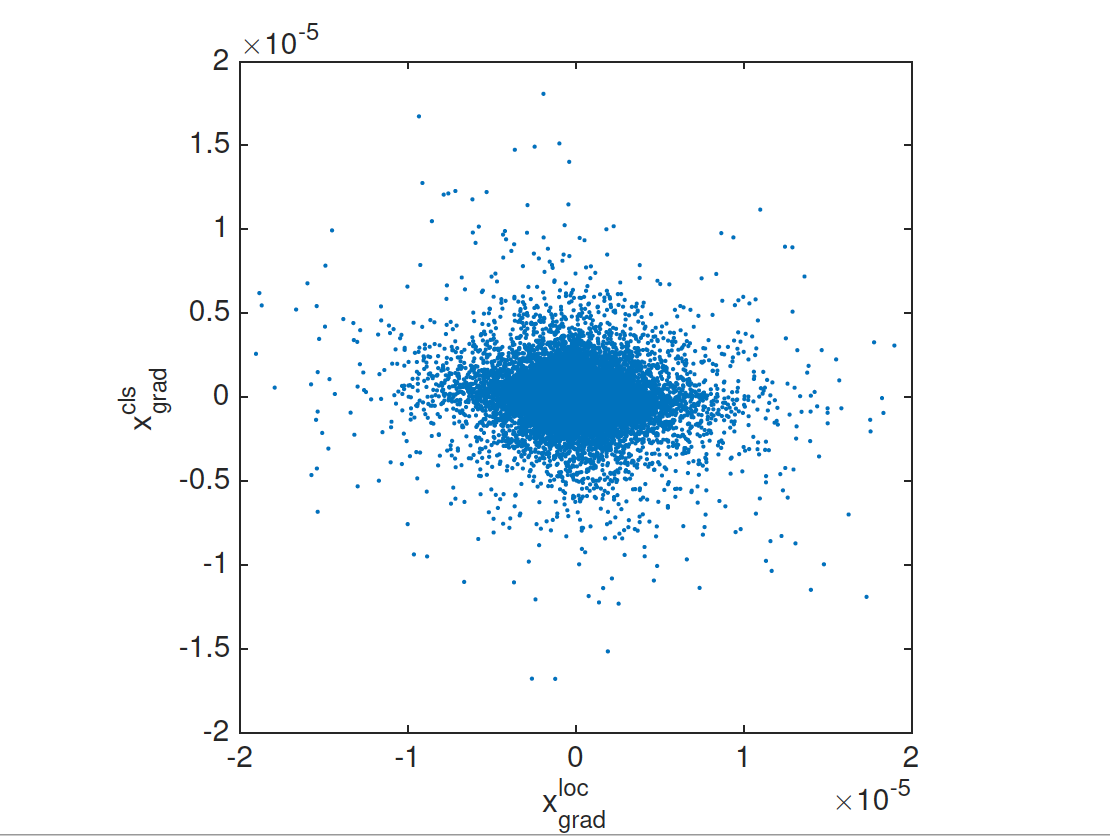}
	\put(8,2){\sffamily\textcolor{black}{{\scalebox{0.7}{\rotatebox{0}{(b)}}}}}	
	\put(50,2){\sffamily\colorbox{white}{\textcolor{black}{{\scalebox{0.8}{\rotatebox{0}{$\g_l$}}}}}}
	\put(5,40){\sffamily\colorbox{white}{\textcolor{black}{{\scalebox{0.8}{\rotatebox{90}{$\g_c$}}}}}}
	\end{overpic}
	\caption{\textbf{Mutual impacts of task losses and  gradient visualization}. (a) Model performance on classification and localization under different attacks: clean image, $\lc$-based attack  and $\ll$-based attack. The model is a standard detector trained on clean images. The performance metric is  detailed in text. (b) Scatter plot of task gradients for classification $\g_c$ and  localization $\g_l$.}
	\label{fig:loss_analysis}
	\vspace{-0.1in}
\end{figure}

\subsection{The Roles of Task Losses in Robustness}\label{sec:task_loss_impacts}
As the classification and localization tasks of a detector share a base-net (\cf Figure~\ref{fig:ssd_chart}), the two tasks will inevitably affect each other even though the input images are  manipulated according to a criterion trailered for one individual task.
We therefore conduct analysis on the role of task losses in model robustness from several perspectives.
{\flushleft \textbf{Mutual Impacts of Task Losses.}}
Our first empirical observation is that 
\emph{different tasks have mutual impacts and the adversarial attacks trailered for one task can reduce the performance   of the model on the other task}.
To show this, we take a marginalized view over one factor while investigating the impact of the other.
For example, when considering classification, we can marginalize out the factor of location and the problem reduces to a multi-label classification task~\cite{multi_label}; on the other hand, when focusing on localization only, we can marginalize out the class information and obtain a class agnostic object detection problem~\cite{selective_search}.
The results with single step PGD and budget 8 are shown in Figure~\ref{fig:loss_analysis} (a).
The performances are measured on detection outputs \emph{prior} to NMS to better reflect the raw performance. 
A candidates set is first determined as the foreground candidates whose prior boxes have an IoU value larger than 0.5 with any of the ground-truth annotation. 
This ensures that   each selected candidate  has a relative clean input both tasks.
For classification, we compute the classification accuracy on the candidate set. For localization, we compute the average IoU of the predicted bounding boxes with ground-truth bounding boxes.
The attack is generated with one-step PGD and a budget of 8.
It can be observed from the results in Figure~\ref{fig:loss_analysis} (a) that the two losses interact with each other.
The attacks based on the classification loss (${\rm loss}_{\rm cls}$) reduces the classification performance and decreases the localization performance at the same time.
Similarly, the localization loss induced attacks  (${\rm loss}_{\rm loc}$) reduces not only the location performance but the classification performance as well.
This can essentially be viewed as a type of cross-task attack transfer: \emph{i.e.}. when using only the classification loss (task) to generate adversarial images,  the attacks can be transferred to localization tasks and reduce its performance and vice versa. 
This is one of the reason why adversarial images generated based on individual task losses (\emph{e.g.}  classification loss~\cite{DAG}) can effectively attack object detectors.

\begin{figure}
	\centering
	\begin{overpic}[viewport=120 100 450 350, clip, height=3cm, width=5cm]{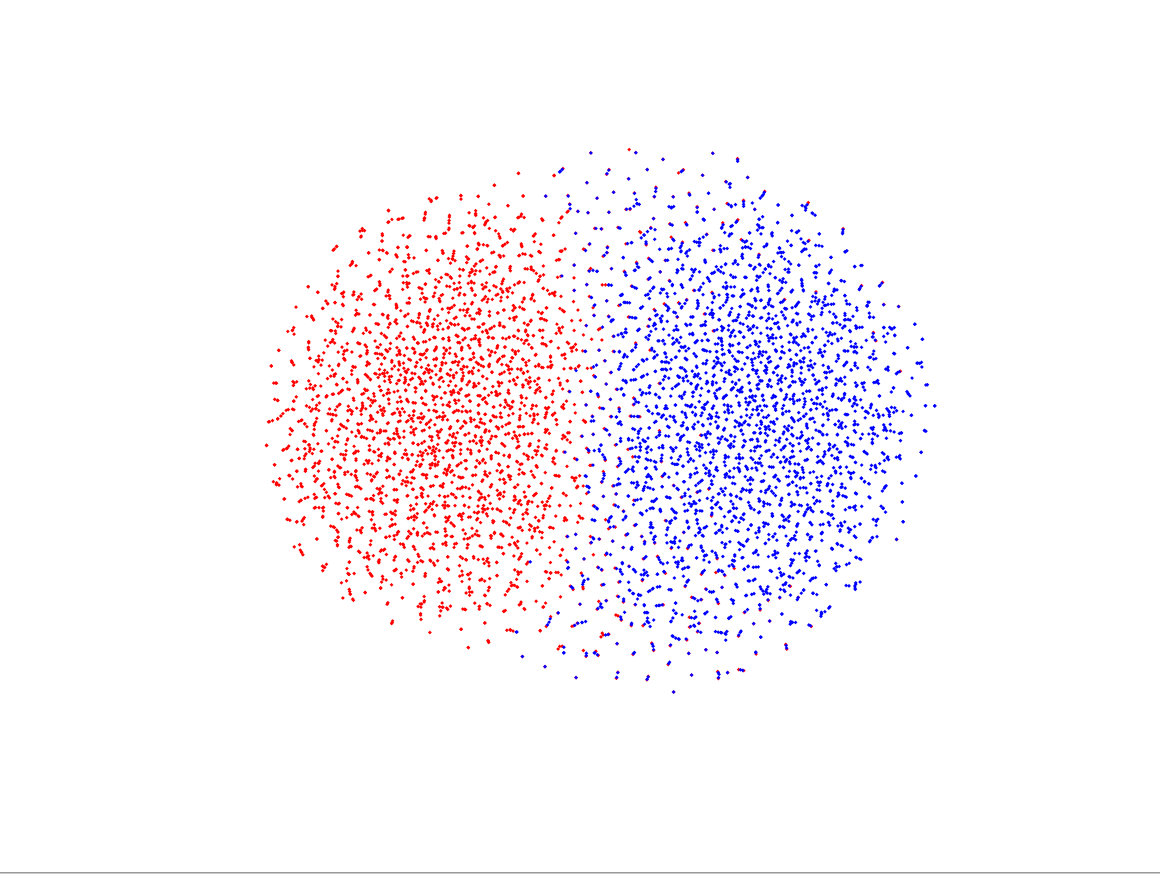}
		\put(98,35){\sffamily\colorbox{white}{\textcolor{black}{{\scalebox{0.8}{\rotatebox{0}{localization}}}}}}
		\put(98,27){\sffamily\colorbox{white}{\textcolor{black}{{\scalebox{0.8}{\rotatebox{0}{task domain}}}}}}
		\put(105,18){\sffamily\colorbox{white}{\textcolor{black}{{\scalebox{0.8}{\rotatebox{0}{$\mathcal{S}_{\rm loc}$}}}}}}
		\put(-30,35){\sffamily{\textcolor{black}{{\scalebox{0.8}{\rotatebox{0}{classification}}}}}}
		\put(-32,27){\sffamily\colorbox{white}{\textcolor{black}{{\scalebox{0.8}{\rotatebox{0}{task domain}}}}}}
		\put(-16,18){\sffamily{\textcolor{black}{{\scalebox{0.8}{\rotatebox{0}{$\mathcal{S}_{\rm cls}$}}}}}}
	\end{overpic}
	\caption{\textbf{Visualization of task domains} $\mathcal{S}_{\rm cls}$ and $\mathcal{S}_{\rm loc}$ using \mbox{t-SNE}. Given a single clean image $\x$, each dot in the picture represents one adversarial example generated by solving Eqn.(\ref{eq:domain}) staring from a random point within the $\epsilon$-ball around $\x$.  Different colors encode the task losses used for generating adversarial examples (\textcolor{red}{red}: $\lc$, \textcolor{blue}{blue}: $\ll$). Therefore, the samples form  empirical images of the corresponding task domains.
	It is observed that the two task domains have both overlaps and distinctive regions.}
	\label{fig:task_domain}
	\vspace{-0.1in}
\end{figure}

%\vspace{-\topsep+0.1in}
{\flushleft \textbf{Misaligned Task Gradients.}}
Our second empirical observation is that \emph{the gradients of the two tasks share certain level of common directions but are not fully aligned, leading to misaligned task gradients that can obfuscate the subsequent adversarial training.}
To show this, 
we analyze the image gradients derived from the two losses (referred to as \emph{task gradients}), \emph{i.e.},
$\g_{c}\!=\!\nabla_{\x}\lc$ and $\g_{l}\!=\!\nabla_{\x}\ll$.
The element-wise scatter plot between $\g_{c}$ and $\g_{l}$ is shown in Figure~\ref{fig:loss_analysis}~(b).
We have several observations: \emph{\textbf{i}}) the magnitudes of the task gradients are not the same (different value ranges), indicating the potential existence of imbalance between the two task losses; \emph{\textbf{ii}}) the direction of the task gradients are inconsistent (non-diagonal), implying the potential conflicts between the two tasks gradients.
We further visualize the task gradient domains representing the domain of a task maximizing gradient for each respective task (\cf Eqn.(\ref{eq:domain}))  as  in Figure~\ref{fig:task_domain}.
The fact the the two domains are not fully separated (\emph{i.e.} they do not collapse to two isolated clusters) further reinforces our previous observation on their mutual impacts. The other aspect that they have a significant non-overlapping portion is another reflection of the mis-alignments between task gradients (task domains).

\subsection{Adversarial Training for Robust Detection}
Motivated by the preceding analysis, we  propose the following formulation for robust object detection training:
\begin{eqnarray}\label{eq:proposed}
	\begin{split}
		\min_{\tt}\big[\max_{\bar{\x} \in \mathcal{S}_{\rm cls} \cup \mathcal{S}_{\rm loc}} \Loss(f_{\tt}(\bar{\x}), \{y_k, \bbox_k\})\big],
	\end{split}
\end{eqnarray}
where the \emph{task-oriented domain} $\mathcal{S}_{\rm cls}$ and 
$\mathcal{S}_{\rm loc}$ represent the permissible domains induced by  each individual tasks: 
\begin{eqnarray}\label{eq:domain}
	\begin{split}
	\mathcal{S}_{\rm cls} &\!\triangleq\! \{\bar{\x} | \, \arg\max_{\bar{\x} \in \mathcal{S}_{\x}} \lc(f(\bar{\x}), \{y_k\}))\} \\
	\mathcal{S}_{\rm loc} &\!\triangleq\! \{\bar{\x} |\, \arg\max_{\bar{\x} \in \mathcal{S}_{\x}} \ll(f(\bar{\x}), \{\bbox_k\})) \}
	\end{split}
\end{eqnarray}
where  $\mathcal{S}_{\x}$ is defined as
$\mathcal{S}_{\x}\!=\!\{\z\,|\, \ \z \!\in\! B(\x, \epsilon) \cap [0, 255]^n\}$,
and $B(\x, \epsilon)=\{\z\,|\,\|\z - \x\|_{\infty} \leq \epsilon\}$ denotes the \mbox{$\ell_{\infty}$-ball} with center as the clean image $\x$ and radius as the perturbation budget $\epsilon$.
We denote $\P_{\mathcal{S}_{\x}}(\cdot)$ as a projection operator projecting the input into the feasible region $\mathcal{S}_{\x}$.
It is important to note several crucial differences compared with the conventional adversarial training for classification:
%\vspace{-\topsep}
\begin{itemize}[leftmargin=0.15in] 
	\item \textbf{multi-task sources for adversary training}: different from the adversarial training in classification case~\cite{FGSM,madry2017towards} where only a single source is involved, here we have \emph{multiple} (in the presence of multiple objects) and \emph{heterogeneous} (both classification and localization) sources of supervisions for adversary generation and training, thus generalizing the adversarial training for classification;
	\item \textbf{task-oriented domain constraints}:
	different from the conventional adversarial training setting which uses a \emph{task-agnostic} domain constraint $\mathcal{S}_{\x}$, 
	we introduce a \emph{task-oriented} domain constraint $\mathcal{S}_{\rm cls}\cup\mathcal{S}_{\rm loc}$ which restricts the permissible domain as the set of  images that maximize either the classification task losses or the localization losses. The final adversarial example used for training is the one that maximizes the overall loss within this set.
	The crucial advantage of the proposed formulation with task-domain constraints is that we can 
	benefit from generating adversarial examples guided by each task without suffering from the interferences between them.
\end{itemize}

\begin{algorithm}[tb]
	\caption{Adversarial Training for Robust Detection}
	\label{alg:MAT}
	\begin{algorithmic}
		\STATE {\bfseries Input:} dataset $\mathcal{D}$, training epochs $T$, batch size $S$,
		\STATE \qquad \quad learning rate $\gamma$, attack budget $\epsilon$
		\FOR{$t=1$ {\bfseries to} $T$}	
		\FOR{ random batch $\{\x^i, \{y^i_k, \bbox_k^i\}\}_{i=1}^S\sim\!\!\mathcal{D}$}
		\STATE $\cdot$ $\tilde{\x}^i \sim  B(\x^i, \epsilon)$
		\STATE  compute attacks in the classification task domain \\
		\STATE  $\cdot$ $\bar{\x}_{\rm cls}^i =  \P_{\mathcal{S}_{\x}}\big(\tilde{\x}^i + \epsilon \cdot \text{sign}\big(\nabla_{\x}\lc(\tilde{\x}^i, \{y^i_k\})\big)\big)$ 
		\STATE compute attacks in the localization task domain \\
		\STATE $\cdot$ $\bar{\x}_{\rm loc}^i \!=\!  \P_{\mathcal{S}_{\x}}\big(\tilde{\x}^i + \epsilon \cdot \text{sign}\big(\nabla_{\x}\ll(\tilde{\x}^i, \{\bbox^i_k\})\big)\big)$ 
		\STATE  compute the final attack examples \\		
		\STATE $\cdot$  $\m = \Loss(\bar{\x}^i_{\rm cls}, \{y^i_k, \bbox^i_k\}) > \Loss(\bar{\x}^i_{\rm loc}, \{y^i_k, \bbox^i_k\})$
		\STATE $\cdot$ $\bar{\x}^i =  \m \odot \bar{\x}_{\rm cls}^i + (1 - \m) \odot \bar{\x}_{\rm loc}^i$ 
		\STATE  perform adversarial  training step \\
		\STATE $\cdot$  $\tt = \tt  - \gamma \cdot \nabla_{\tt}\frac{1}{S}\sum_{i=1}^S\Loss(\bar{\x}^i, \{y^i_k, \bbox^i_k\};\tt)$
		\ENDFOR
		\ENDFOR
		\STATE {\bfseries Output:} learned model parameter $\tt$ for object detection.
	\end{algorithmic}
\end{algorithm}

If we relax the task-oriented domain to $\mathcal{S}_{\x}$,  set the coordinates of the bounding box corresponding to the full image and assign a single class label to the image, then  the proposed formulation Eqn.(\ref{eq:proposed}) reduces to the conventional adversarial training setting for classification~\cite{FGSM,madry2017towards}. Therefore, we can view the proposed adversarial training for robust detection as a natural generalization of the conventional adversarial training under the classification setting.
However, it is crucial to note that while both tasks contribute to improving the model robustness in expectation according to their overall strengths,  there is no interference between the tasks for generating individual adversarial example due to the task oriented domain  in contrast to $\mathcal{S}_{\x}$ (\cf Sec.\ref{sec:task_domain}).

Training  object detection models that are resistant to adversarial attacks boils down to solving a minimax problem as in  Eqn.(\ref{eq:proposed}).
We solve  it approximately by replacing the original  training images with the adversarially perturbed ones obtained by solving the inner problem, and then conducting  conventional training of the model using the perturbed images as typically done in  adversarial training~\cite{FGSM,madry2017towards}. 
The inner maximization is approximately solved using a variant of FGSM~\cite{FGSM}  for efficiency.
For incorporating the task-oriented domain constraint, we propose to take FGSM steps within each task domain and then select the one that maximizes the overall loss.
The details of the  algorithm are summarized in Algorithm~\ref{alg:MAT}.

\begin{figure}
	\begin{overpic}[viewport=5 2 350 195, clip, width=4cm]{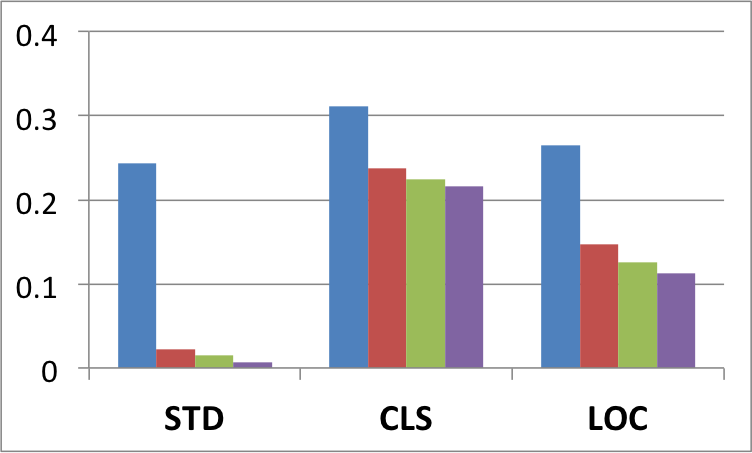}
		\put(5,1){\sffamily\textcolor{black}{{\scalebox{0.6}{\rotatebox{0}{(a)}}}}}
		%\put(35,-5){\sffamily\textcolor{black}{{\scalebox{0.6}{attack steps ($\lc$)}}}}
		\put(-5,25){\sffamily\textcolor{black}{{\scalebox{0.6}{\rotatebox{90}{mAP}}}}}
	\end{overpic}
	\hspace{0.07in}
	\begin{overpic}[viewport=5 2 350 180, clip, width=4cm]{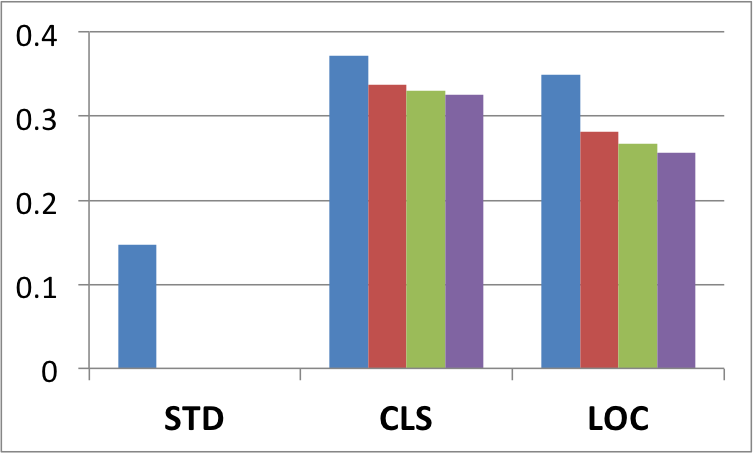}
		\put(-35,52){\includegraphics[viewport=100 180 260 200, clip, width=2.5cm]{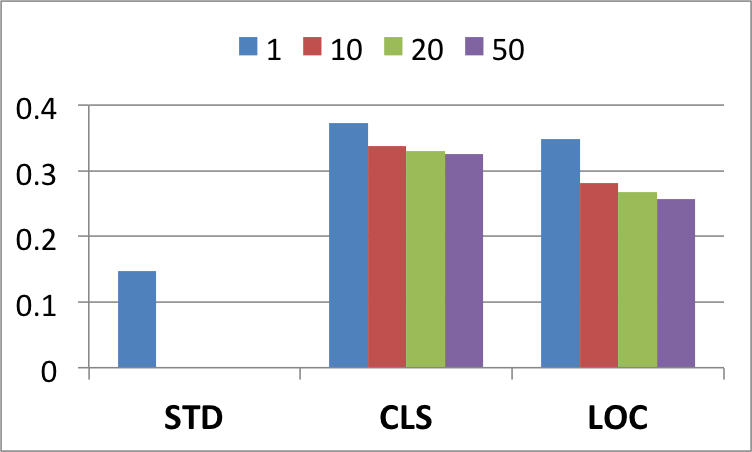}}
		\put(6,1){\sffamily\textcolor{black}{{\scalebox{0.6}{\rotatebox{0}{(b)}}}}}
		%\put(35,-5){\sffamily\textcolor{black}{{\scalebox{0.6}{attack steps ($\ll$)}}}}
		\put(-5,25){\sffamily\textcolor{black}{{\scalebox{0.6}{\rotatebox{90}{mAP}}}}}
	\end{overpic}
	\caption{\textbf{Model performance under different number of  steps} for  (a)~$\lc$ and (b) $\ll$-based PGD attack  with $\epsilon=8$. \salg{STD} is the standard model. \salg{CLS} and \salg{LOC} are our robust models.}
	\label{fig:attack_steps}
	\vspace{-0.2in}
\end{figure}

\section{Experiments}

\subsection{Experiment and Implementation  Details}

We use the single-shot multi-box detector (SSD)~\cite{ssd} with VGG16~\cite{VGG} backbone as one of the representative single-shot detectors in our experiments.
We also make the necessary modifications to the VGG16 net as detailed in~\cite{ssd} and keep the batch normalization layers.
Experiments with different detector architectures (Receptive Field Block-based Detector (RFB)~\cite{RFB}, Feature Fusion Single Shot Detector (FSSD)~\cite{FSSD} and YOLO-V3~\cite{yolo, YOLOv3}) and backbones (VGG16~\cite{VGG}, ResNet50~\cite{resnet}, DarkNet53~\cite{darknet13}) are also conducted for comprehensive evaluations.

For PASCAL VOC dataset, we adopt the standard ``07+12'' protocol (a union of 2007 and 2012 \texttt{trainval}, $\mathtt{\sim}$16k images) following~\cite{ssd} for training. For testing, we use PASCAL VOC2007 \texttt{test} with 4952 test images and 20 classes~\cite{VOC}.\footnote{VOC2012 \texttt{test} is not used as the annotations required for generating attacks are unavailable.}
For MS-COCO dataset~\cite{COCO}, we train on \texttt{train+valminusminival} 2014 ($\mathtt{\sim}$120k images) and test on \texttt{minival} 2014 with 80 classes ($\mathtt{\sim}$5k images) .
The ``mean average precision'' (mAP)  with IoU threshold 0.5 is used for evaluating the performance of a detector~\cite{VOC}.

All models are trained from scratch using SGD with an initial learning rate of $10^{-2}$, momentum $0.9$, weight decay $0.0005$ and batch size $32$~\cite{pretrain} with the multi-box loss~\cite{multibox_cvpr,multibox2}.
The learning rate schedule is [40k, 60k, 80k] for PASCAL VOC and [180k, 220k, 260k] for MS-COCO  with decay factor 0.1.
The size of the image is $300 \!\times\!300$.
Pixel value range is $[0, 255]$ shifted according to dataset mean.
For adversarial attacks and training, we use a budget $\epsilon=8$, which roughly corresponds to a PSNR of 30 between the perturbed and original images following~\cite{BMVC_RAP}. All the attack methods incorporate  $\sgn(\cdot)$ operator into the PGD steps  for normalization and efficiency following~\cite{FGSM}.

\subsection{Impacts of Task Losses on Robustness}
We will investigate the role of task losses in model robustness.
For this purpose, we introduce the standard  model and several  variations of our proposed robust model:
\begin{itemize}[leftmargin=0.15in]
	\item \salg{STD}: standard training with clean image as the domain
	\item \salg{CLS}: using $\mathcal{S}_{\rm cls}$  only as the task domain for training
	\item \salg{LOC}: using $\mathcal{S}_{\rm loc}$  only as the task domain for training.
\end{itemize}
We will systemically investigate the performance of these models under attacks induced by \emph{individual task losses} with different number of attack steps and budgets as follows.

\begin{figure}[t]
	\begin{overpic}[viewport=5 2 350 195, clip, width=4cm]{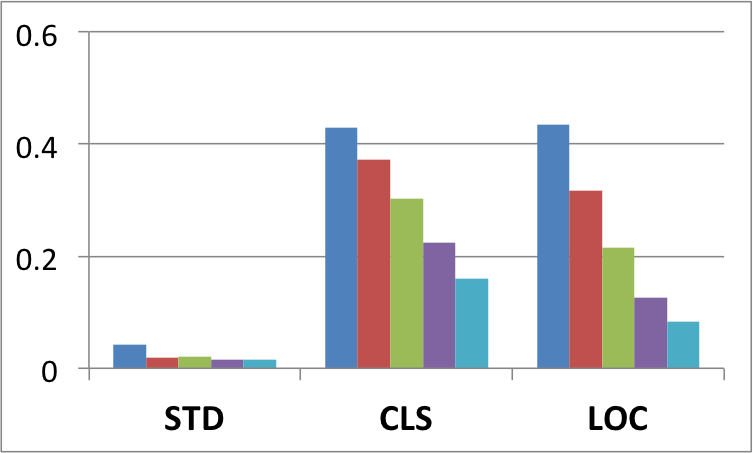}
		\put(5,1){\sffamily\textcolor{black}{{\scalebox{0.6}{\rotatebox{0}{(a)}}}}}
		%\put(35,-5){\sffamily\textcolor{black}{{\scalebox{0.6}{attack steps ($\lc$)}}}}
		\put(-5,25){\sffamily\textcolor{black}{{\scalebox{0.6}{\rotatebox{90}{mAP}}}}}
	\end{overpic}
	\hspace{0.07in}
	\begin{overpic}[viewport=5 2 350 180, clip, width=4cm]{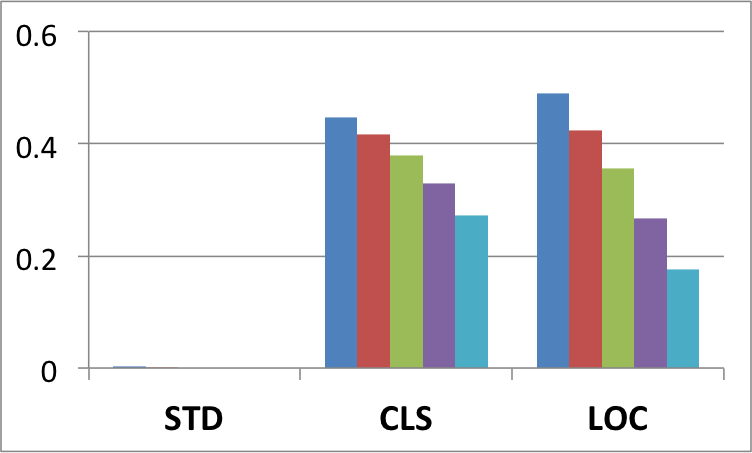}
		\put(-40,52){\includegraphics[viewport=90 180 280 200, clip, width=3.cm]{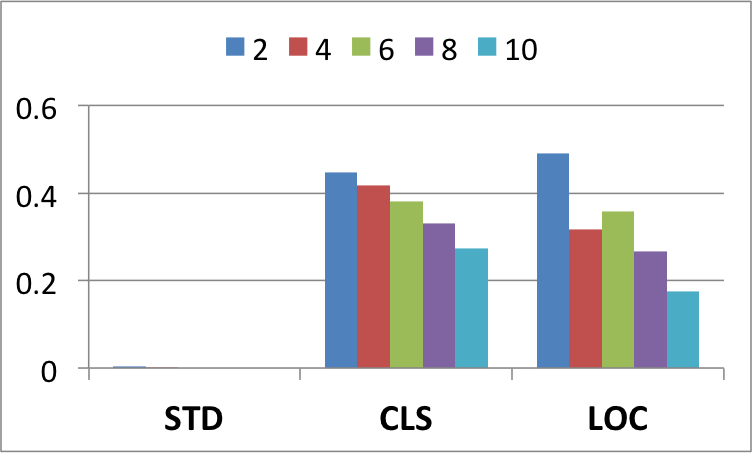}}
		\put(6,1){\sffamily\textcolor{black}{{\scalebox{0.6}{\rotatebox{0}{(b)}}}}}
		%\put(35,-5){\sffamily\textcolor{black}{{\scalebox{0.6}{attack steps ($\ll$)}}}}
		\put(-5,25){\sffamily\textcolor{black}{{\scalebox{0.6}{\rotatebox{90}{mAP}}}}}
	\end{overpic}
	%\vspace{0.01in}
	\caption{\textbf{Model performance under different attack budgets} for  (a)~$\lc$ and (b) $\ll$-based PGD attack with 20 steps. \salg{STD} is the standard model. \salg{CLS} and \salg{LOC} are our robust models.}
	\label{fig:attack_budget}
\end{figure}

\begin{figure}
	\centering
	\begin{overpic}[viewport=160 60 640 555, clip, width=2cm]{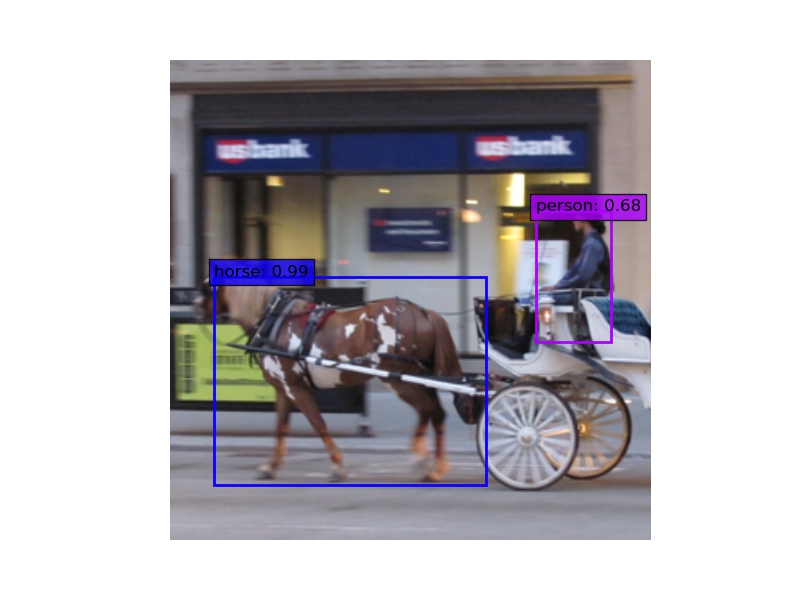}
		\put(40,-9){\sffamily\textcolor{black}{{\scalebox{0.7}{\rotatebox{0}{$\epsilon=0$}}}}}
	\end{overpic}
	\begin{overpic}[viewport=160 60 640 555, clip, width=2cm]{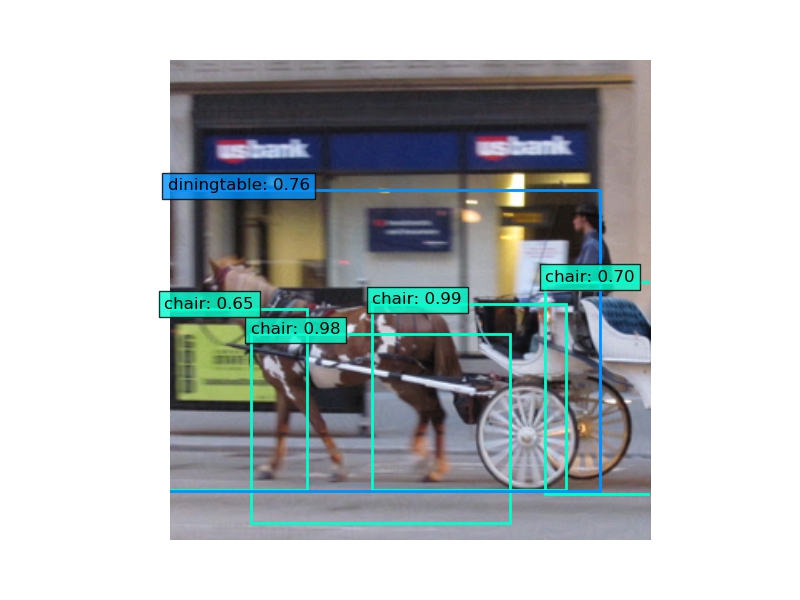}
		\put(40,-9){\sffamily\textcolor{black}{{\scalebox{0.7}{\rotatebox{0}{$\epsilon=2$}}}}}
	\end{overpic}
	\begin{overpic}[viewport=160 60 640 555, clip, width=2cm]{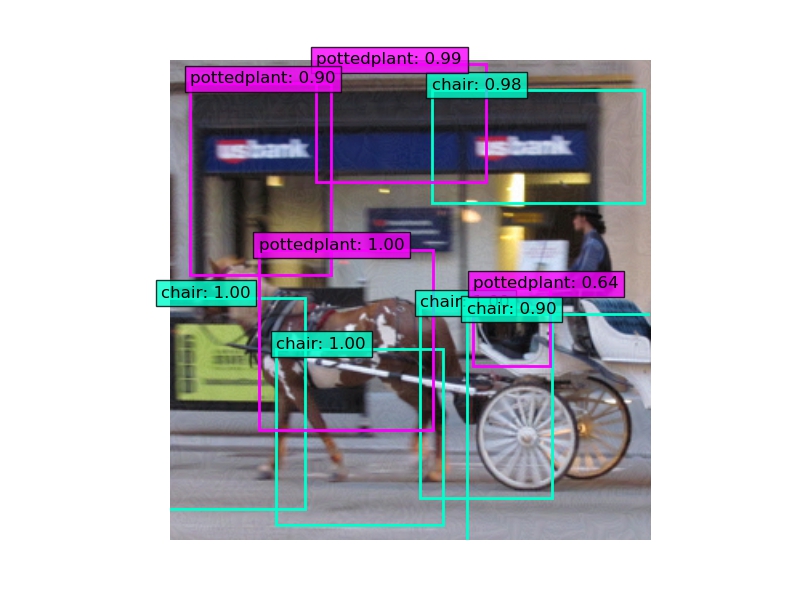}
		\put(40,-9){\sffamily\textcolor{black}{{\scalebox{0.7}{\rotatebox{0}{$\epsilon=4$}}}}}
	\end{overpic}
	\begin{overpic}[viewport=160 60 640 555, clip, width=2cm]{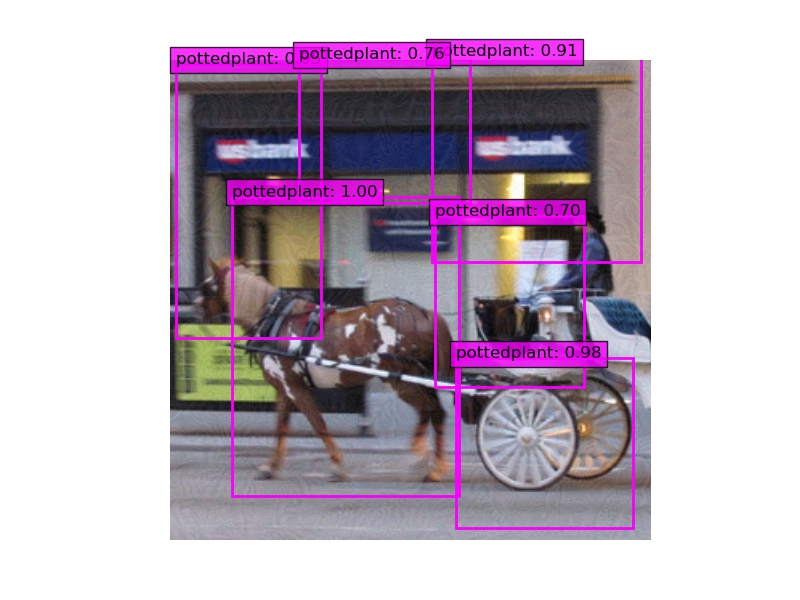}
		\put(40,-9){\sffamily\textcolor{black}{{\scalebox{0.7}{\rotatebox{0}{$\epsilon=8$}}}}}
	\end{overpic}
	\vspace{0.05in}
	\caption{\textbf{Visualization of attacks} on \salg{STD} model using $\lc$ based 20-step PGD attack (zoom electronically for better view).}
	\label{fig:viz_budget}
	\vspace{-0.1in}
\end{figure}
\vspace{-\topsep}
{\flushleft \textbf{Attacks under different number of steps.}} We first evaluate the performance of models under attacks with different number of PGD steps under a fixed attack budget of 8. The results are shown in Figure~\ref{fig:attack_steps}.
We have several interesting observations from the results:
\emph{\textbf{i}})~the performance of the standard model (\salg{STD}) 
drops below all other robust models within just a few steps and
 decreases quickly (approaching zero) as the number of PGD steps increases, for both $\lc$-base and $\ll$-based attacks.
These results imply that both types of attacks  are very effective attacks for detectors;
\emph{\textbf{ii}})~all the robust models maintains a relative stable performance across different number of attack steps, indicating their improved robustness against adversarial attacks compared to the standard model.
\vspace{-\topsep}
{\flushleft \textbf{Attacks with different budgets.}}
We evaluate  model robustness under a range of different attack budgets $\epsilon \in \{2, 4, 6, 8, 10\}$. 
The results are presented in Figure~\ref{fig:attack_budget}.
It is observed that the performance of the standard model trained with natural images (\salg{STD}) drops significantly, \emph{e.g.}, from $\mathtt{\sim}72\%$ on clean images (not shown in figure) to  $\mathtt{\sim}4\%$ with a small attack budget of 2.
Robust models, on the other hand, degrade more gracefully as the attack budget increases, implying their improved robustness compared to the standard model.
In Figure~\ref{fig:viz_budget}, we visualize the detection results under different attack budgets on standard model.
It is observed that even with a small attack budget (\emph{e.g.} $\epsilon\!=\!2$), the detection results are changed completely, implying that the standard model is very fragile in term of robustness, which is consistent with our previous observation from  Figure~\ref{fig:attack_budget}.
It is also observed that the erroneous detections can be of several forms: \emph{\textbf{i}}) label flipping: the bounding box location is roughly correct but the class label is incorrect, \emph{e.g.}, ``\name{dinningtable}''  ($\epsilon: 0 \!\rightarrow \! 2$);
\emph{\textbf{ii}}) disappearing: the bounding box for the object is missing, \emph{e.g.}, ``\name{horse}'' and ``\name{person}'' ($\epsilon: 0 \!\rightarrow\! 2$);
\emph{\textbf{iii}}) appearing: spurious detections of objects that  do not exist in the image with locations not well aligned with any of the dominant objects,  \emph{e.g.},  ``\name{chair}'' ($\epsilon: 0\! \rightarrow\! 2$) and ``\name{pottedplant}''   ($\epsilon: 2\! \rightarrow\! 8$).
As the attack budget is increased, the detection output will be further changed in terms of the three types of changes  described above. 
It can also be observed from the figure that the attack image generated with $\epsilon=8$ bears noticeable changes compared with the original one, although  not very severe.  
We will therefore use attack $\epsilon=8$ as it is a large enough attack budget while maintain a reasonable resemblance to the original image.

\begin{table}[t]
	\centering
	\begin{tabular}{P{0.7cm}V{1}M{0.5cm}|M{0.7cm}V{1} R{0.8cm} R{0.8cm} V{1}  R{0.9cm} R{0.9cm} }
		\multicolumn{2}{c|}{ attacks} & clean & $\lc$ & $\ll$   & \salg{DAG}~\cite{DAG} & \salg{RAP}~\cite{BMVC_RAP}   \\
		\Xhline{2\arrayrulewidth}
		\multicolumn{2}{c|}{\salg{standard}} & 72.1 & 1.5  & 0.0  &	0.3 & 6.6 \\
		\Xhline{1\arrayrulewidth}
		\multirow{4}{*}{\salg{ours}} &\salg{CLS}       & 
		46.7&	21.8& 32.2  &	28.0& 43.4 \\
		& \salg{LOC}     &  51.9&	23.7&	26.5  &  17.2 & 43.6\\
		& \salg{CON}     & 38.7 & 18.3 & 27.2&  26.4 & 40.8 \\
		& \salg{MTD} &  48.0 & 29.1  & 31.9 &	28.5	& 44.9\\
		\Xhline{0.1\arrayrulewidth}
		\multicolumn{2}{c|}{\salg{ours avg}} & 46.3  &  23.2  &  29.4  &  25.0 &    43.2
	\end{tabular}
	\caption{Impacts of task domains on model performance (mAP) and defense against attacks from literature (attack $\epsilon=8$). }
	\label{tab:task_domain}
	\vspace{-0.1in}
\end{table}

\subsection{Beyond Single-Task Domain}\label{sec:task_domain}
We  further examine  the impacts of task domains on  robustness.
The following  approaches with different task domains are considered in addition to \alg{STD}, \alg{CLS} and \alg{LOC}:
\begin{itemize}[leftmargin=0.15in] %\itemsep-0.05in 
	\item \salg{CON}: using the conventional task agnostic domain $\mathcal{S}_{\x}$, which is essentially the direct application of the adversarial training for classification~\cite{FGSM, madry2017towards} to detection; 
	\item \salg{MTD}: using the task oriented domain $\mathcal{S}_{\rm cls} \cup  \mathcal{S}_{\rm loc}$.
\end{itemize}
The results are summarized in Table~\ref{tab:task_domain}.
It is observed from comparison
 that different domains lead to different levels of model robustness.
For example, for methods with a single task domain,  \salg{LOC}  leads to less robust models compared with \salg{CLS}. On the other hand, \salg{LOC} has a higher clean accuracy than \salg{CLS}. 
Therefore, it is not straightforward to select one single domain as it is \emph{unknown a priori} whether one of the task domains is the best.
Simply relaxing the task domains as done in the conventional adversarial training \salg{CON}~\cite{FGSM, madry2017towards} leads to compromised performance.
Concretely, the performance of \salg{CON} with task-agnostic task domain achieves an in-between or inferior  performance compared to the models with individual task domains under different  attacks, implying that simply mixing the task domains leads to compromised performance, due to the conflicts between the task gradients (Sec.~\ref{sec:task_loss_impacts}). 
On the other hand, the robust model \salg{MTD}   using adversarial training with  task oriented domain constraint  can improve the performance over  \salg{CON} baseline.
More importantly,  when  the task-oriented multi-task domain is incorporated, a proper trade-off and overall performance is observed compared with the single domain-based methods, implying the importance of properly handling  heterogeneous and possibly imbalanced tasks in object detectors.
In summary, the tasks could be imbalanced and contribute differently to the model robustness. As it is \emph{unknown a priori} which is better, randomly adopting one or simply combining the losses (\alg{CON}) could lead to compromised performance. \salg{MTD} setting overcomes this issue and achieves performance on par or better than best single domain models and  the task-agnostic domain model.

\begin{figure*}[t]
	\centering
	\begin{overpic}[viewport=160 60 640 555, clip, width=3.3cm]{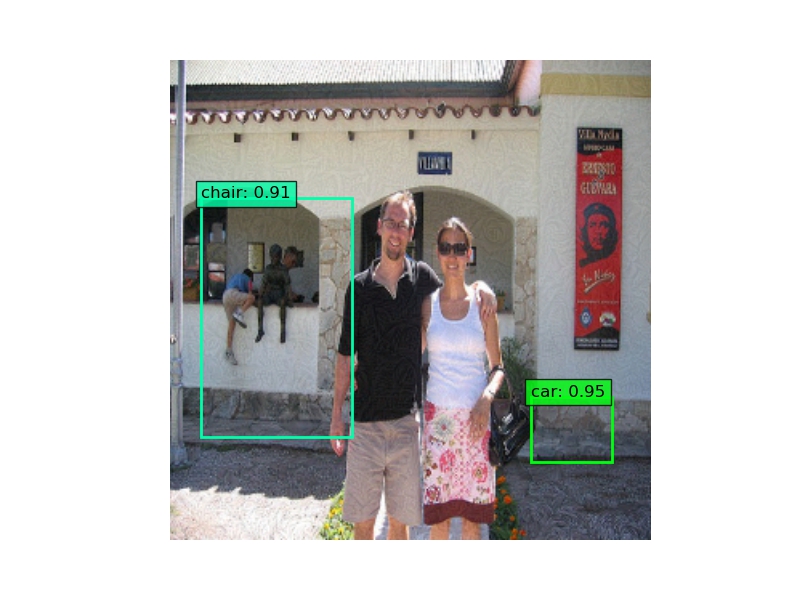}
		\put(-7,25){\sffamily\textcolor{black}{{\scalebox{0.8}{\rotatebox{90}{standard}}}}}
	\put(180,100){\sffamily\textcolor{black}{{\scalebox{0.8}{\rotatebox{0}{Results under the DAG~\cite{DAG} attack}}}}}
	\end{overpic}
	\begin{overpic}[viewport=160 60 640 555, clip, width=3.3cm]{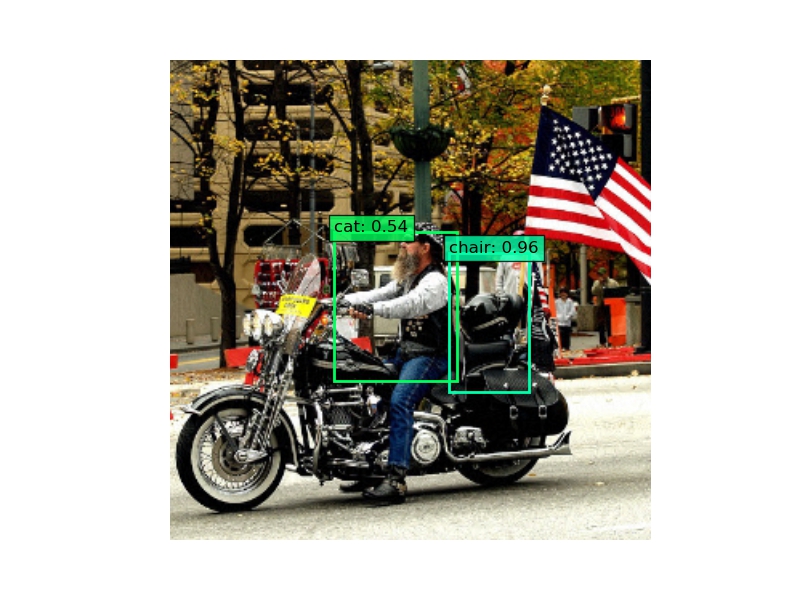}
	\end{overpic}
	\begin{overpic}[viewport=160 60 640 555, clip, width=3.3cm]{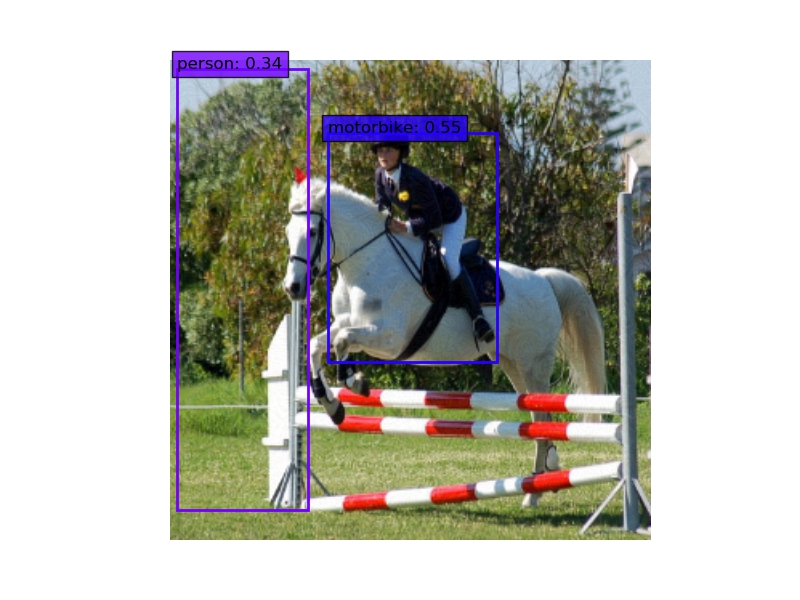}
	\end{overpic}
	\begin{overpic}[viewport=160 60 640 555, clip, width=3.3cm]{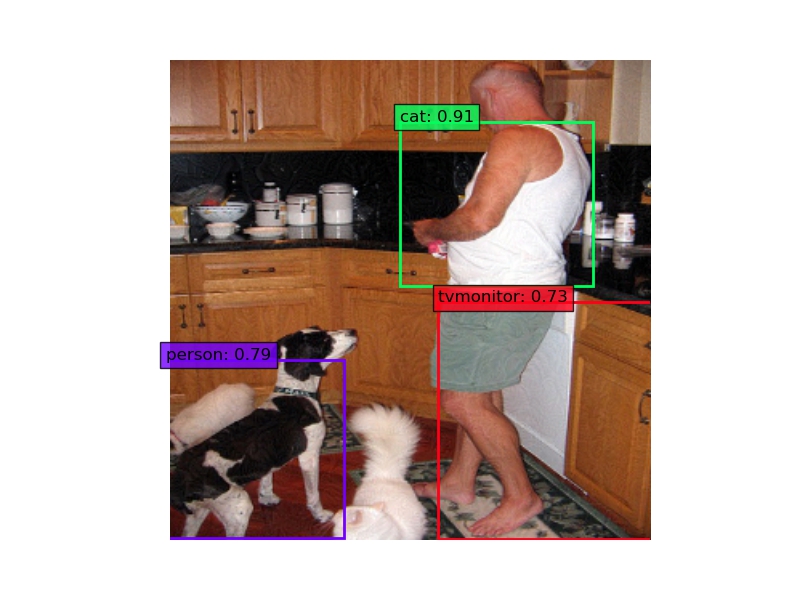}
	\end{overpic}
	\begin{overpic}[viewport=160 60 640 555, clip, width=3.3cm]{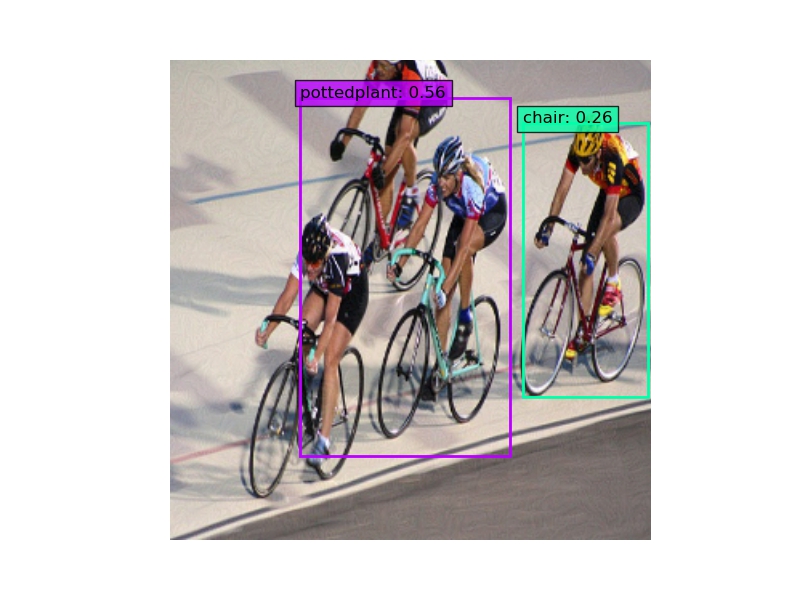}
	\end{overpic}
	%---------
	\begin{overpic}[viewport=160 60 640 555, clip, width=3.3cm]{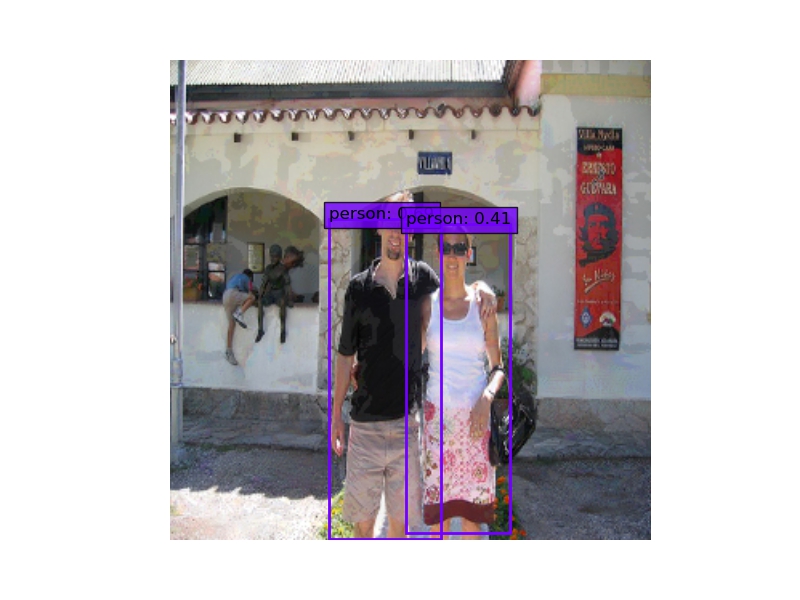}
		\put(-7,40){\sffamily\textcolor{black}{{\scalebox{0.8}{\rotatebox{90}{ours}}}}}
	    \put(-4,-3.5){\dashline{3}(0,0)(500,0)}
	\end{overpic}
	\begin{overpic}[viewport=160 60 640 555, clip, width=3.3cm]{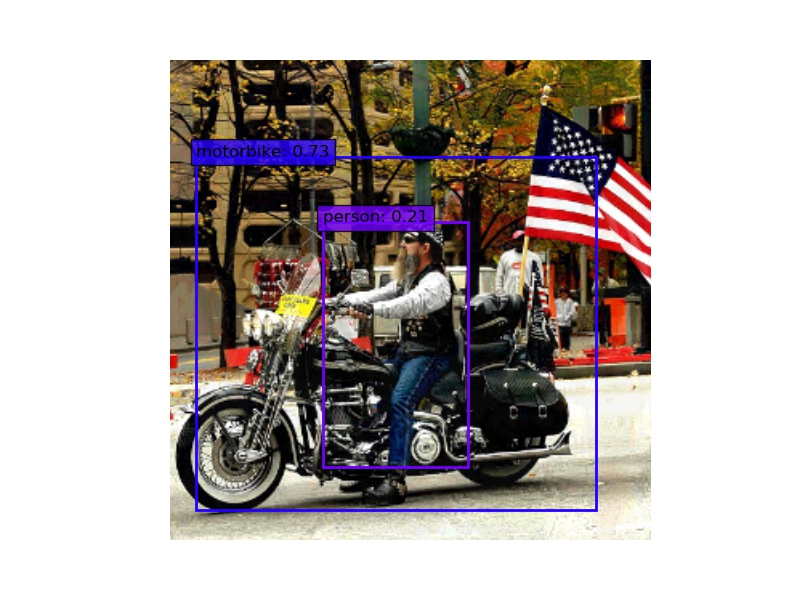}
	\end{overpic}
	\begin{overpic}[viewport=160 60 640 555, clip, width=3.3cm]{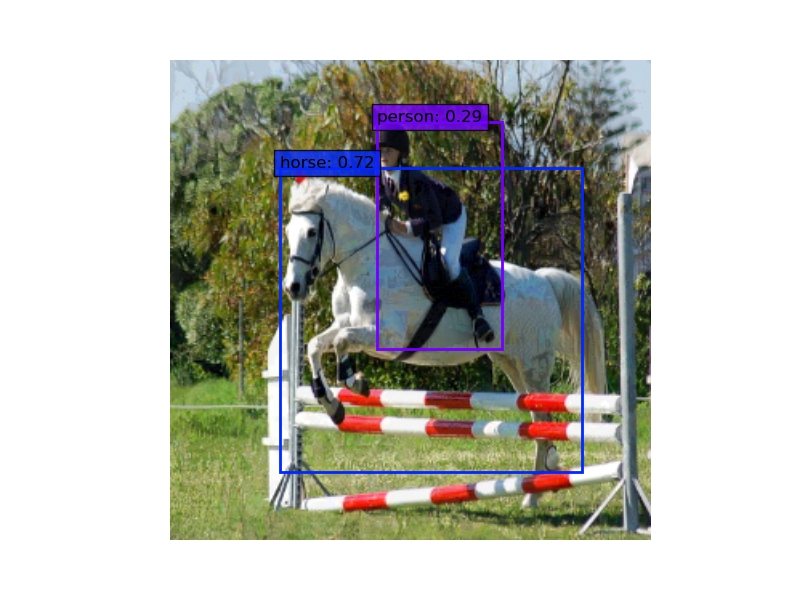}
	\end{overpic}
	\begin{overpic}[viewport=160 60 640 555, clip, width=3.3cm]{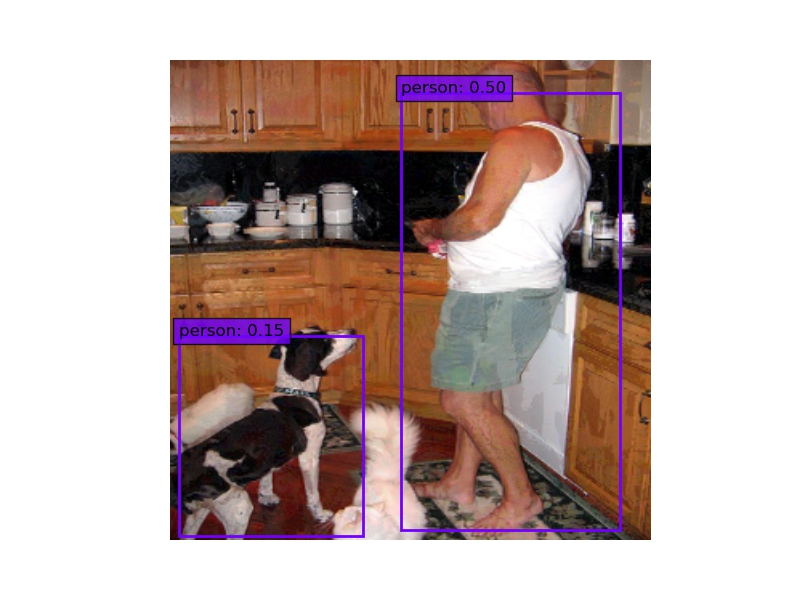}
	\end{overpic}
	\begin{overpic}[viewport=160 60 640 555, clip, width=3.3cm]{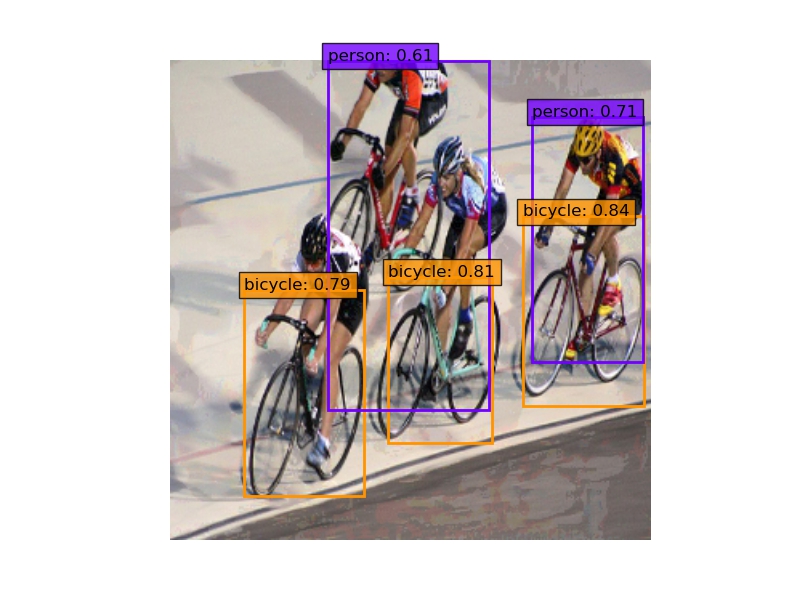}
	\end{overpic}\\
	\vspace{0.05in}
	%---------
	\begin{overpic}[viewport=160 60 640 555, clip, width=3.3cm]{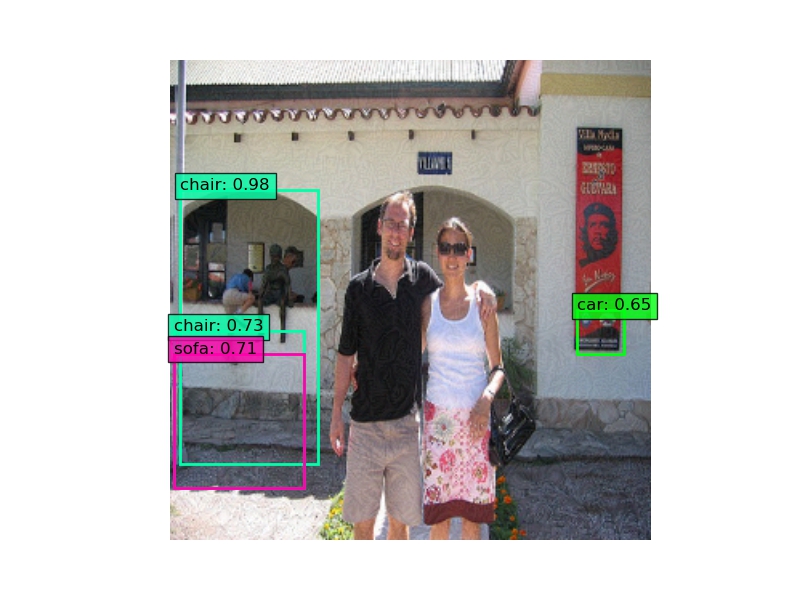}
		\put(-7,25){\sffamily\textcolor{black}{{\scalebox{0.8}{\rotatebox{90}{standard}}}}}
	\end{overpic}
	\begin{overpic}[viewport=160 60 640 555, clip, width=3.3cm]{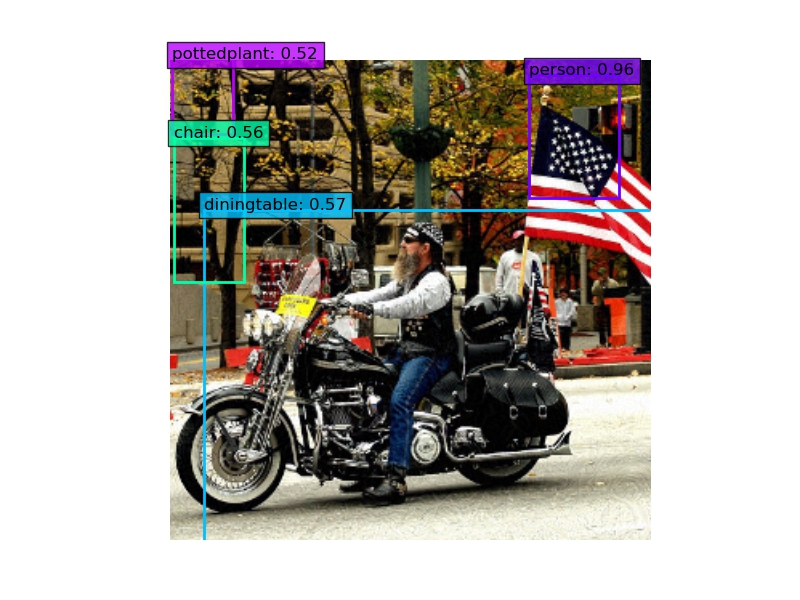}
	\end{overpic}
	\begin{overpic}[viewport=160 60 640 555, clip, width=3.3cm]{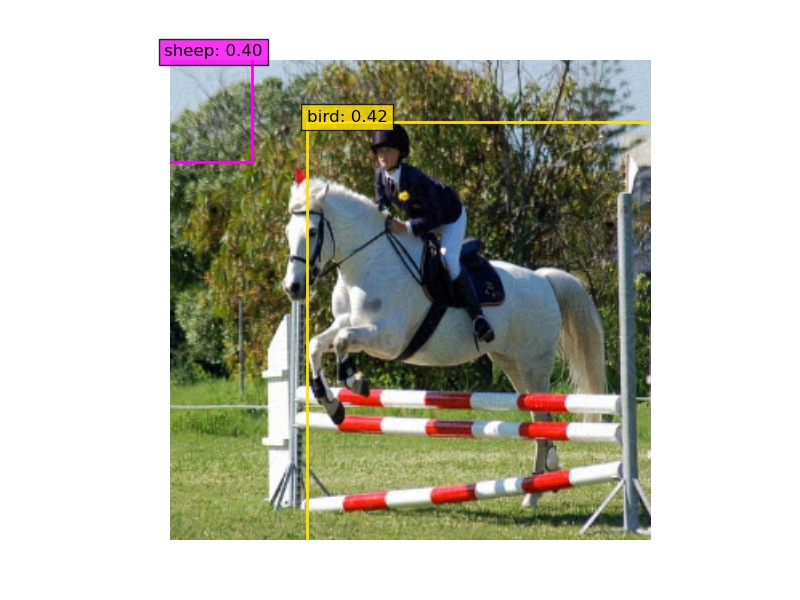}
	\end{overpic}
	\begin{overpic}[viewport=160 60 640 555, clip, width=3.3cm]{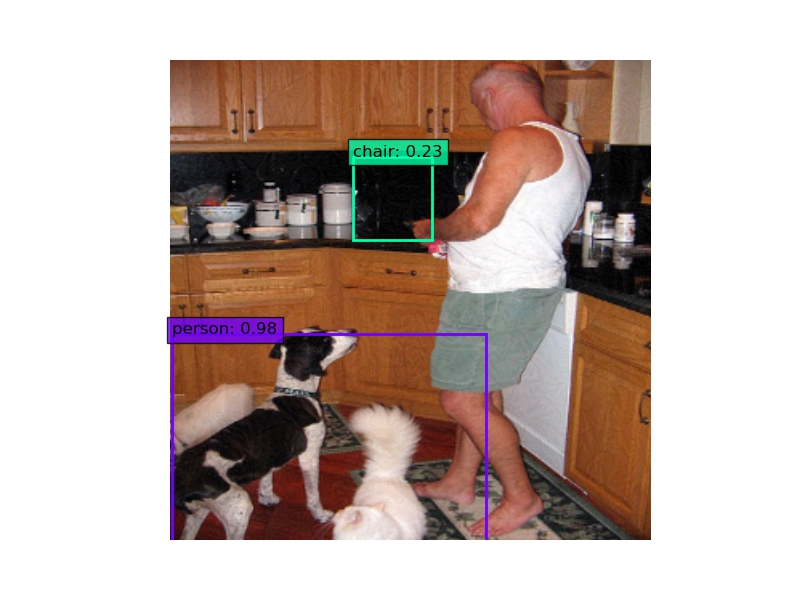}
	\end{overpic}
	\begin{overpic}[viewport=160 60 640 555, clip, width=3.3cm]{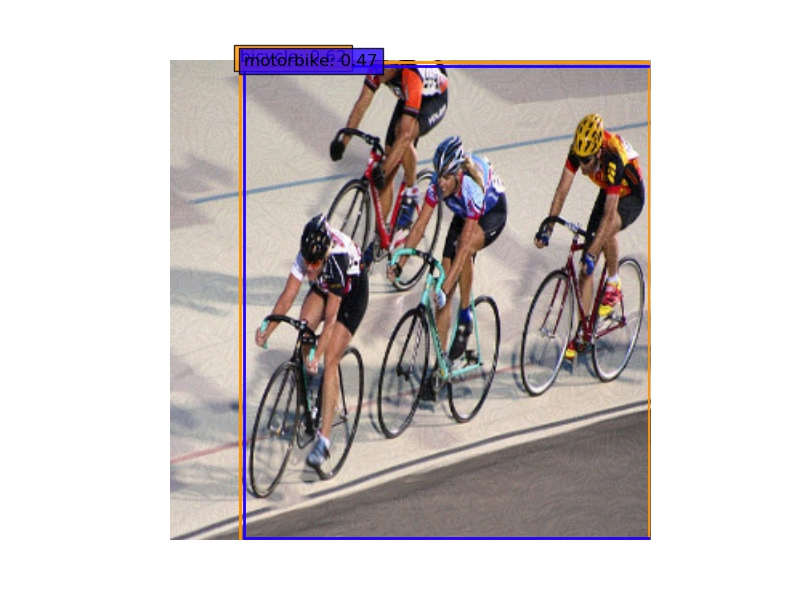}
	\end{overpic}
	%---------
	\begin{overpic}[viewport=160 60 640 555, clip, width=3.3cm]{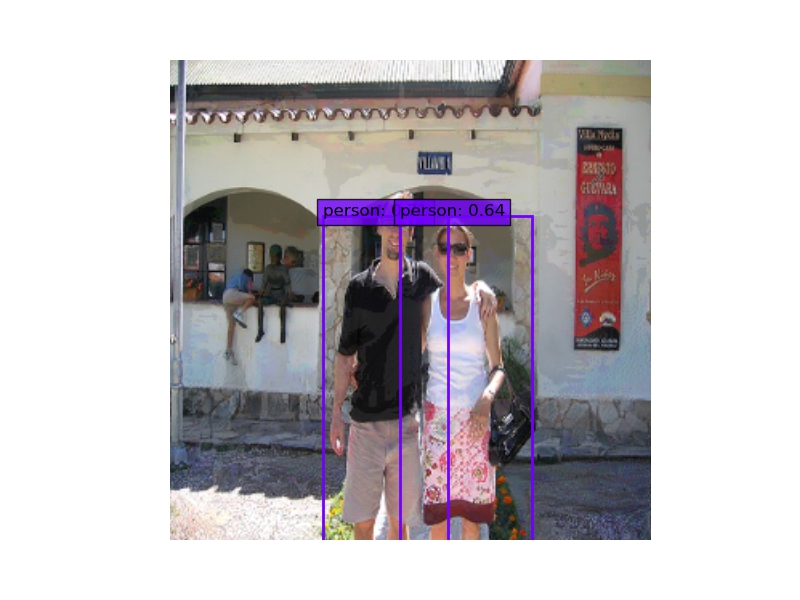}
		\put(-7,40){\sffamily\textcolor{black}{{\scalebox{0.8}{\rotatebox{90}{ours}}}}}
		\put(180,-10){\sffamily\textcolor{black}{{\scalebox{0.8}{\rotatebox{0}{Results under the RAP~\cite{BMVC_RAP} attack}}}}}
	\end{overpic}
	\begin{overpic}[viewport=160 60 640 555, clip, width=3.3cm]{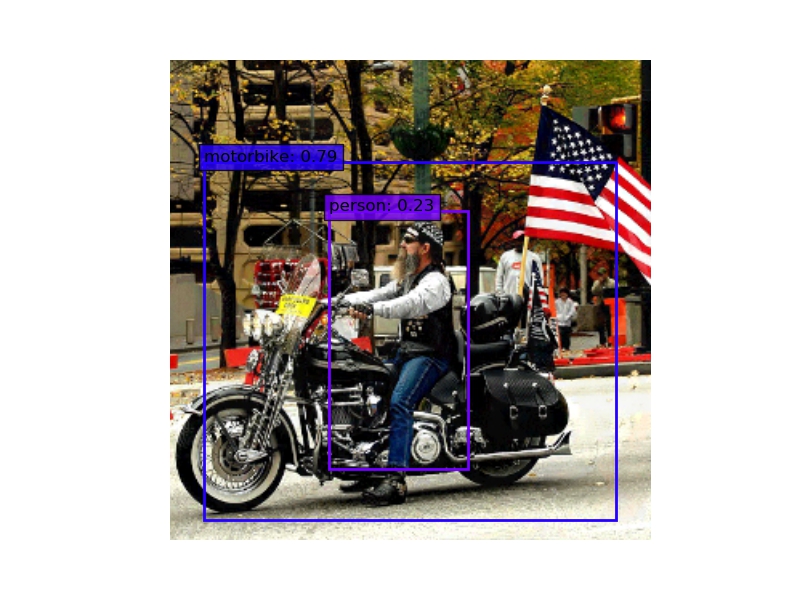}
	\end{overpic}
	\begin{overpic}[viewport=160 60 640 555, clip, width=3.3cm]{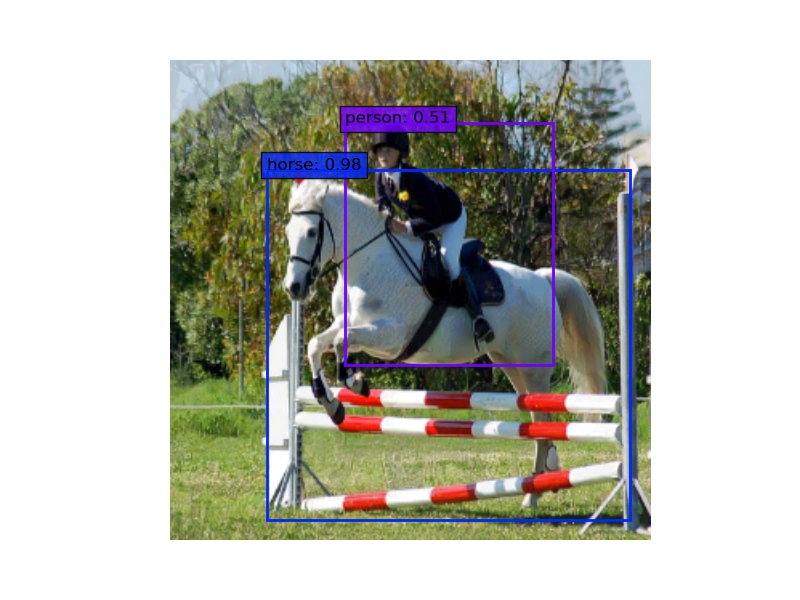}
	\end{overpic}
	\begin{overpic}[viewport=160 60 640 555, clip, width=3.3cm]{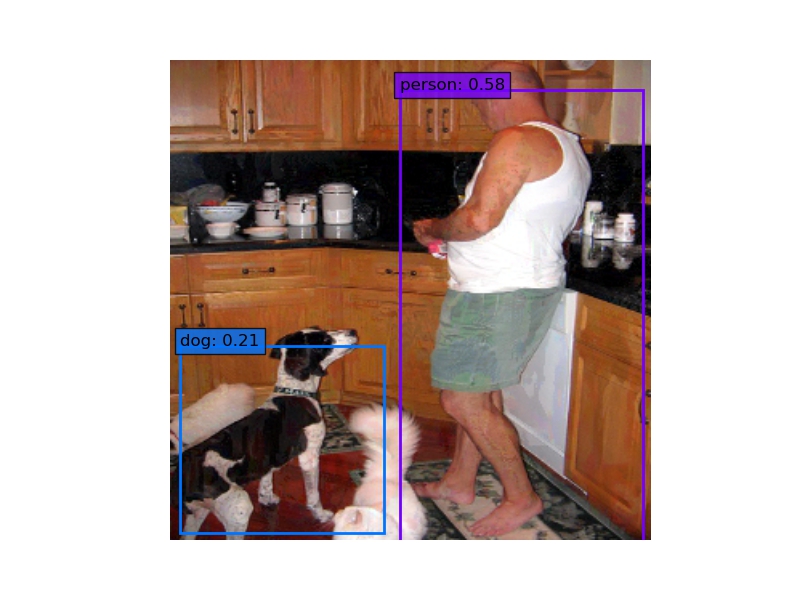}
	\end{overpic}
	\begin{overpic}[viewport=160 60 640 555, clip, width=3.3cm]{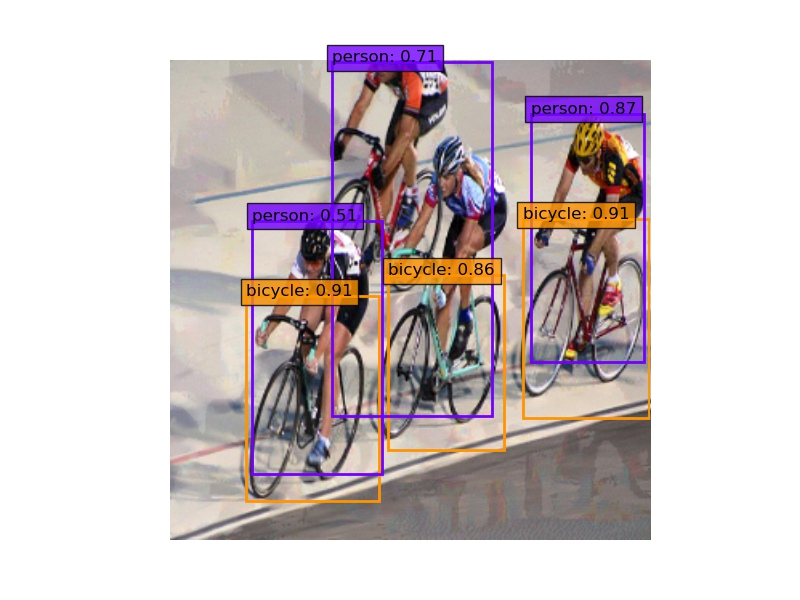}
	\end{overpic}
	\vspace{0.1in}
	\caption{\textbf{Visual comparison} between \salg{standard} model and \salg{ours} under DAG~\cite{DAG} and  RAP~\cite{BMVC_RAP} attacks with  attack budget 8.}
	\label{fig:viz_results}
	\vspace{-0.15in}
\end{figure*}

\subsection{Defense against Existing White-box Attacks}
To further investigate the model robustness, we evaluate models against representative attack methods from literature.
We use  \alg{DAG}~\cite{DAG} and \alg{RAP}~\cite{BMVC_RAP} as  representative attacks  according to Table~\ref{tab:attack_analysis}. It is important to note that the attack used in training and testing are different.
The results are summarized in Table~\ref{tab:task_domain}.
It is observed that the performances of robust models improve over the standard model by a large margin.
\alg{CLS}  performs better in general than \alg{LOC} and \alg{CON} in terms of robustness against the two attacks from literature.
The model using multi-task domains (\alg{MTD}) demonstrates the best performance.
\alg{MTD}  has a higher clean image accuracy than \alg{CLS} and performs uniformly well against different attacks,  thus overall is better
and will be used  for reporting performance in the following.
Visualization of  example results are provided in Figure~\ref{fig:viz_results}.

\begin{table}[t]
	\centering
	\begin{tabular}{M{2.cm}|M{0.7cm}|M{0.7cm}|M{0.7cm}|M{0.7cm}}
		\rule{0pt}{0.05pt}\multirow{2}{*}{\small {SSD-backbone}} & \multicolumn{2}{c|}{\salg{DAG}~\cite{DAG}}& \multicolumn{2}{c}{\salg{RAP}~\cite{BMVC_RAP}} \\[-0.8ex]
		\rule{0pt}{0.05pt} &\salg{STD} & \salg{ours} & \salg{STD} & \salg{ours}   \\[-0.5ex]
		\Xhline{2\arrayrulewidth}
		{\small  VGG16}  &  0.3 &	  28.5 & 6.6 & 44.9\\
		{\small  ResNet50}  & 0.4 &   22.9 &  8.8 & 39.1\\
		{\small  DarkNet53} & 0.5 &   26.2 &  8.2 & 46.6 \\
	\end{tabular}
	\caption{Evaluation  results on  across different backbones. }
	\label{tab:backbones}
	\vspace{-0.15in}
\end{table}

\subsection{Evaluation on Different Backbones}
We evaluate the effectiveness of the proposed approach under different SSD backbones, including VGG16~\cite{VGG}, ResNet50~\cite{resnet} and DarkNet53~\cite{darknet13}.
Average performance under  \alg{DAG}~\cite{DAG} and \alg{RAP}~\cite{BMVC_RAP} attacks are reported in \mbox{Table~\ref{tab:backbones}}.
It is observed that the proposed approach can boost the performance of the detector by a large margin (20\%$\mathtt{\sim}$30\% absolute improvements), across different backbones, demonstrating that the proposed approach performs well across backbones of different network structures with clear and consistent improvements over baseline models.

\begin{table}[t]
	\centering
	\begin{tabular}{p{0.7cm} p{1.25cm}|M{0.7cm}|M{0.7cm}|M{0.7cm}|M{0.7cm}}
		 \multicolumn{2}{c|}{\multirow{2}{*}{architecture}}   & \multicolumn{2}{c|}{\salg{DAG}~\cite{DAG}}& \multicolumn{2}{c}{\salg{RAP}~\cite{BMVC_RAP}} \\[-0.8ex]
		\rule{0pt}{0.001pt}&&\salg{STD} & \salg{ours} & \salg{STD} & \salg{ours}   \\ [-0.5ex]
		\Xhline{2\arrayrulewidth}
		{\small SSD}  &{\scriptsize+VGG16}  &  0.3 &28.5 & 6.6 & 44.9\\
		{\small RFB}& {\scriptsize+ResNet50} & 0.4 &  27.4 & 8.7& 48.7 \\
		{\small FSSD} &{\scriptsize+DarkNet53} & 0.3&  29.4 & 7.6& 46.8 \\
		{\small YOLO} &{\scriptsize+DarkNet53}  & 0.1 &  27.6& 8.1& 44.3\\
	\end{tabular}
	\caption{Evaluation results on  different detection architectures. }
	\label{tab:archtect}
	\vspace{-0.18in}
\end{table}

\subsection{Results on Different Detection Architectures}
Our proposed approach is also applicable to   different detection architectures.
To show this, we use different detection architectures, including SSD~\cite{ssd}, RFB~\cite{RFB}, FSSD~\cite{FSSD} and YOLO-V3~\cite{yolo, YOLOv3}. 
The input image size for YOLO is $416\!\times \! 416$ and all others take $300\!\times \! 300$ images as input.
Average performance under  \alg{DAG}~\cite{DAG} and \alg{RAP}~\cite{BMVC_RAP} attacks  are summarized in Table~\ref{tab:archtect}.
It is observed  that the proposed method can improve over the standard method significantly and consistently for different detector architectures.
This  clearly demonstrates the applicability of  the proposed approach across detector  architectures.

\begin{figure*}[h]
	\centering
	\begin{overpic}[viewport=160 60 640 555, clip, width=2.1cm]{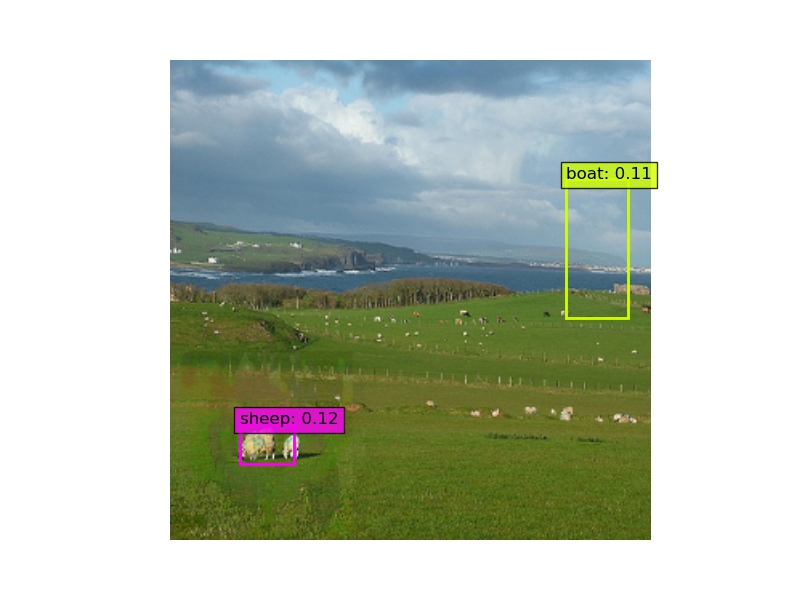}
	\end{overpic}
	\begin{overpic}[viewport=160 60 640 555, clip, width=2.1cm]{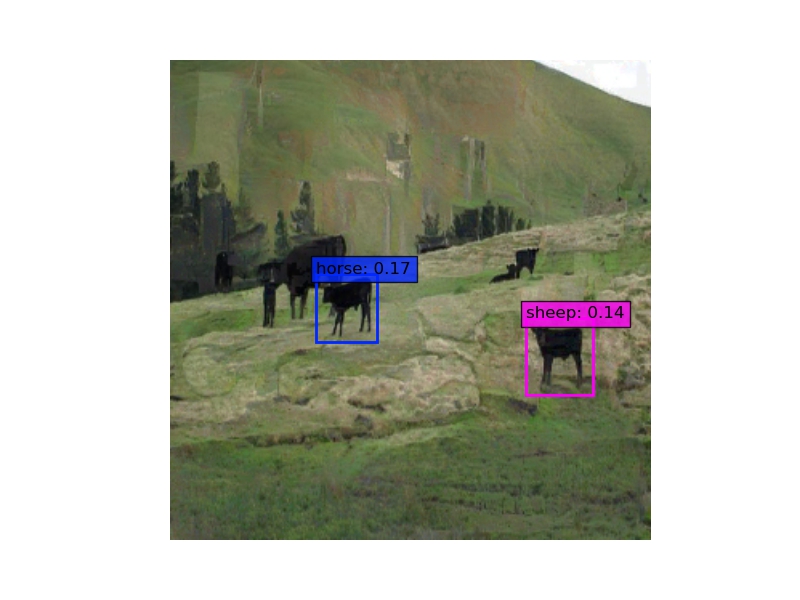}
	\end{overpic}
	\begin{overpic}[viewport=160 60 640 555, clip, width=2.1cm]{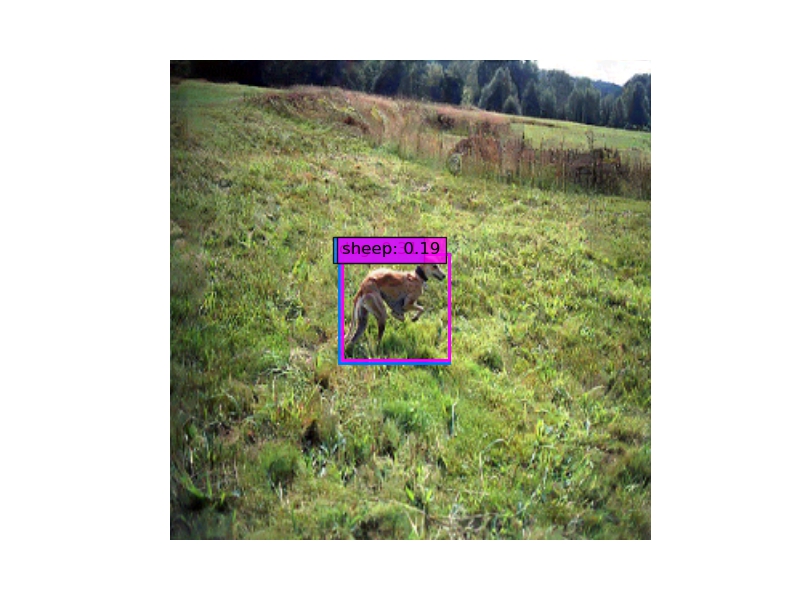}
	\end{overpic}
	\begin{overpic}[viewport=160 60 640 555, clip, width=2.1cm]{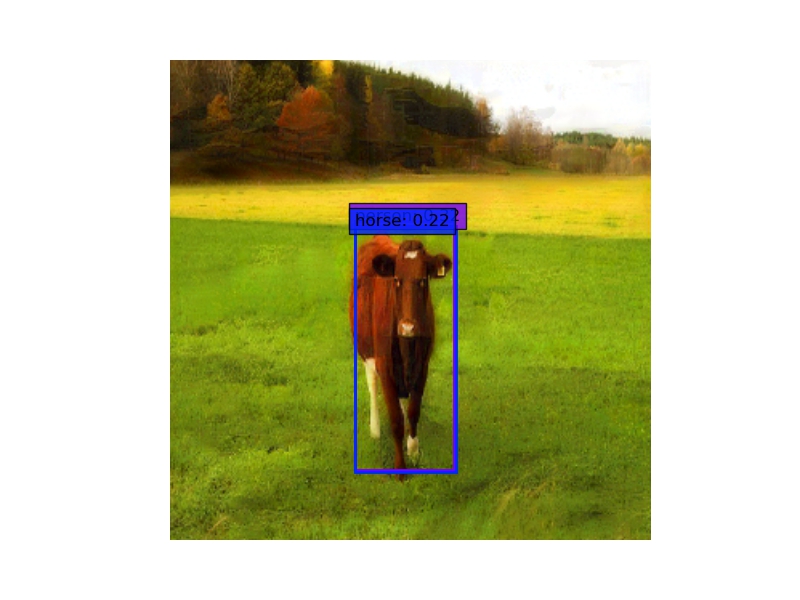}
	\end{overpic}
	\begin{overpic}[viewport=160 60 640 555, clip, width=2.1cm]{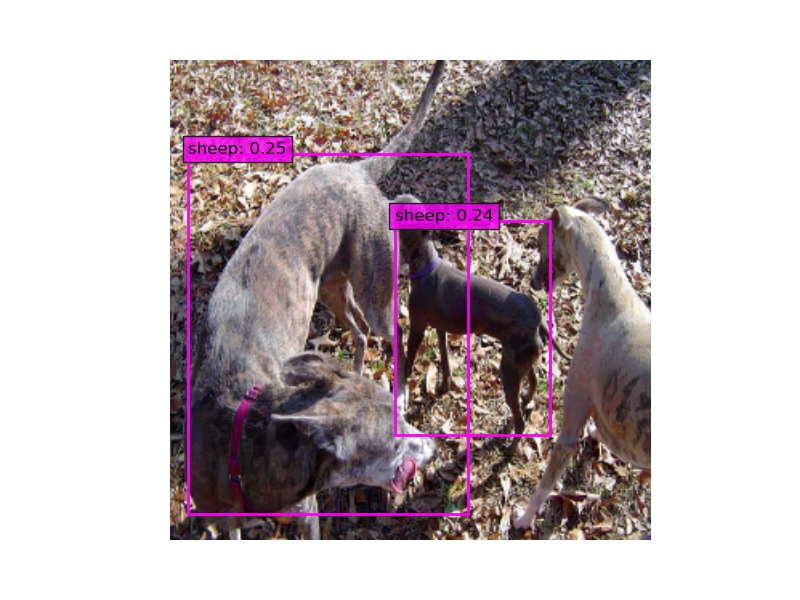}
	\end{overpic}
	\begin{overpic}[viewport=160 60 640 555, clip, width=2.1cm]{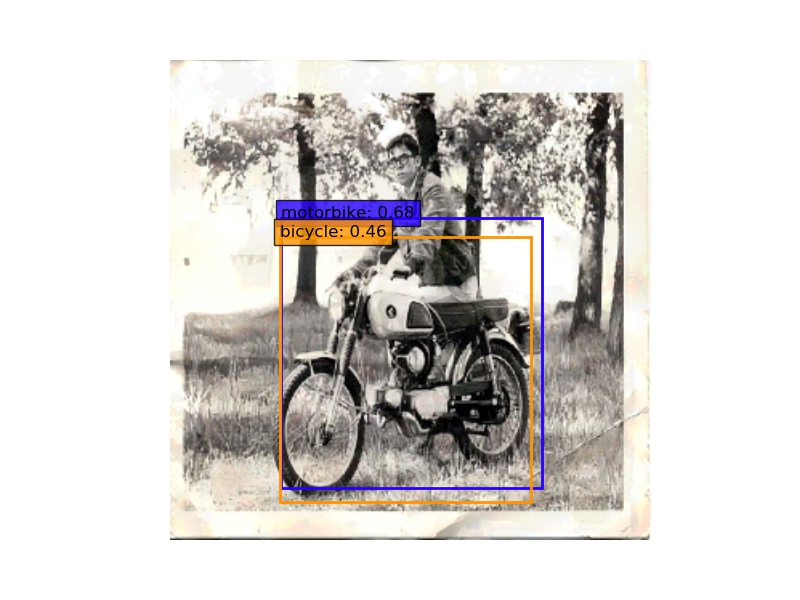}
	\end{overpic}
	\begin{overpic}[viewport=160 60 640 555, clip, width=2.1cm]{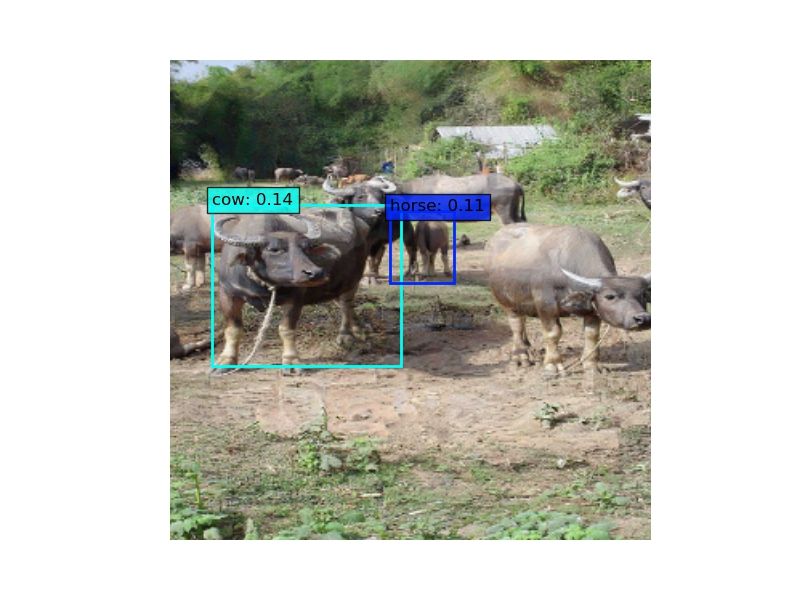}
	\end{overpic}
	\begin{overpic}[viewport=160 60 640 555, clip, width=2.1cm]{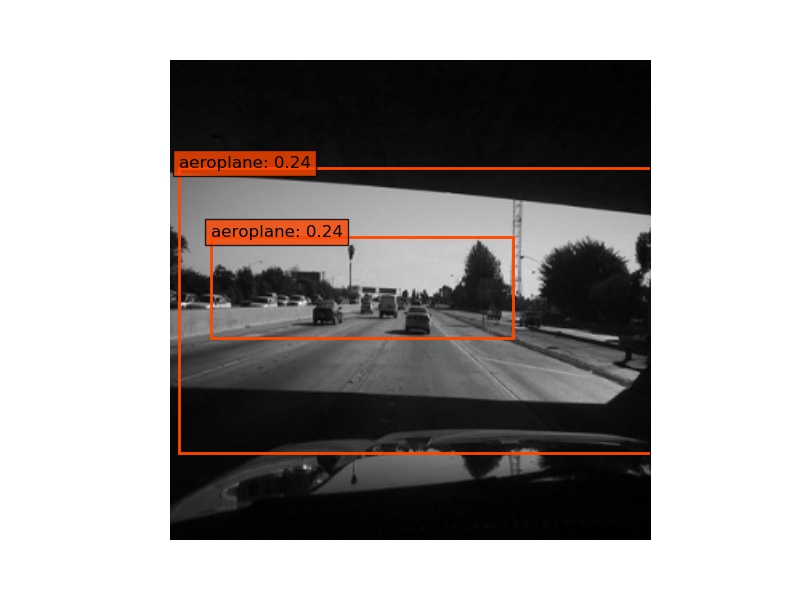}
	\end{overpic}
	\begin{overpic}[viewport=160 60 640 555, clip, width=2.1cm]{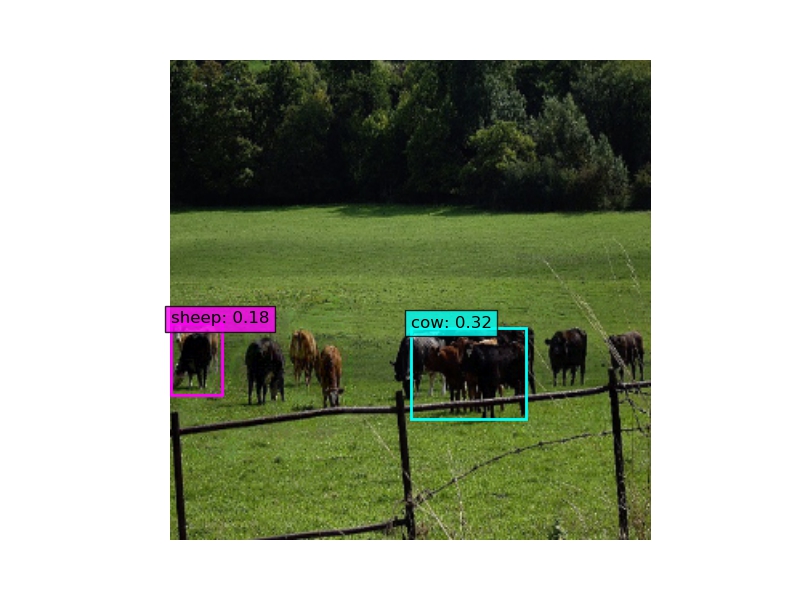}
	\put(110,205){\sffamily\textcolor{black}{{\scalebox{0.75}{\rotatebox{0}{small objects}}}}}
	\end{overpic}
	\begin{overpic}[viewport=160 60 640 555, clip, width=2.1cm]{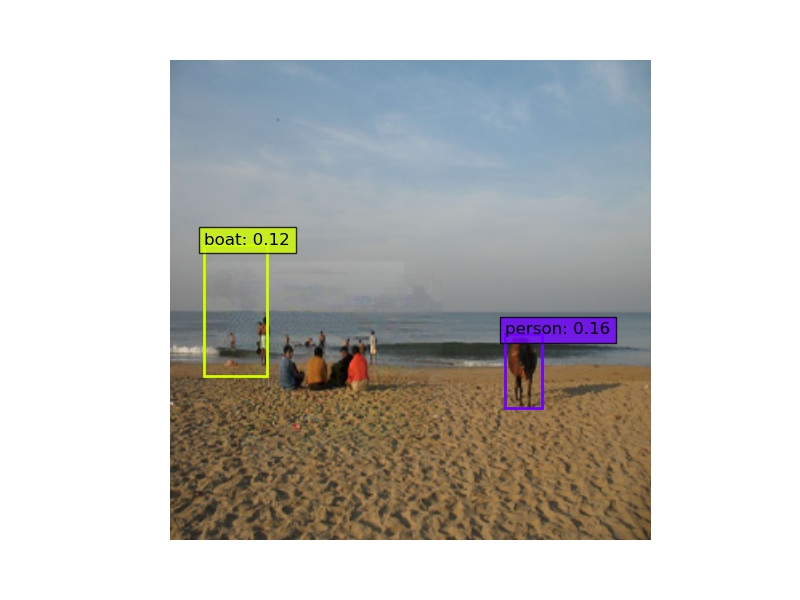}
	\end{overpic}
	\begin{overpic}[viewport=160 60 640 555, clip, width=2.1cm]{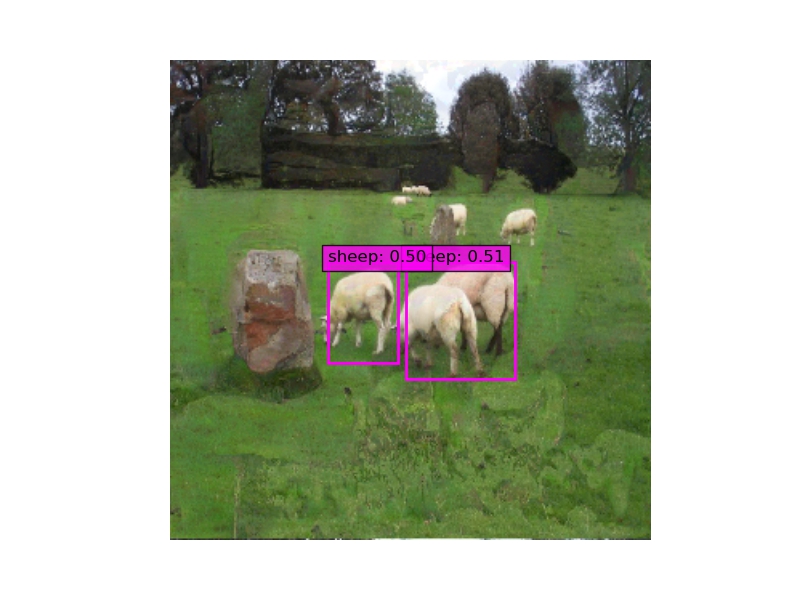}
	\put(130,205){\sffamily\textcolor{black}{{\scalebox{0.75}{\rotatebox{0}{visually confusing classes}}}}}
	\end{overpic}
	\begin{overpic}[viewport=160 60 640 555, clip, width=2.1cm]{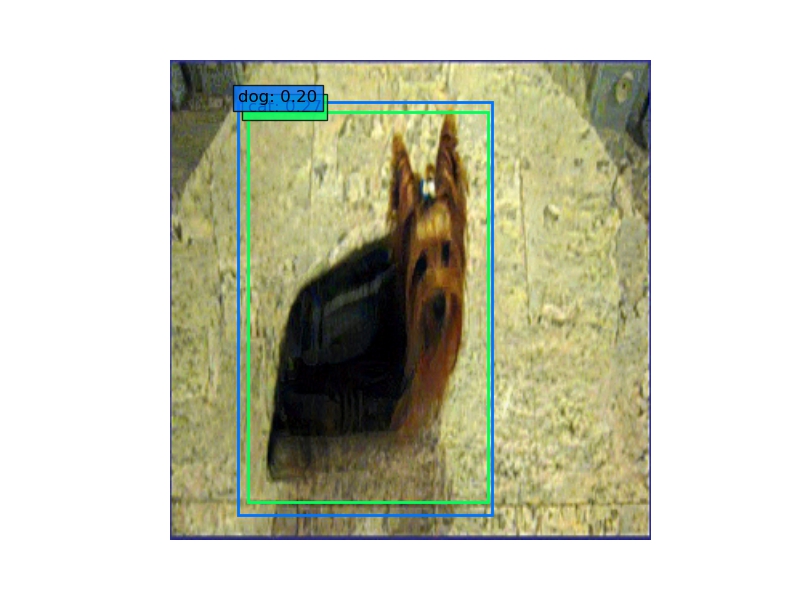}
	\end{overpic}
	\begin{overpic}[viewport=160 60 640 555, clip, width=2.1cm]{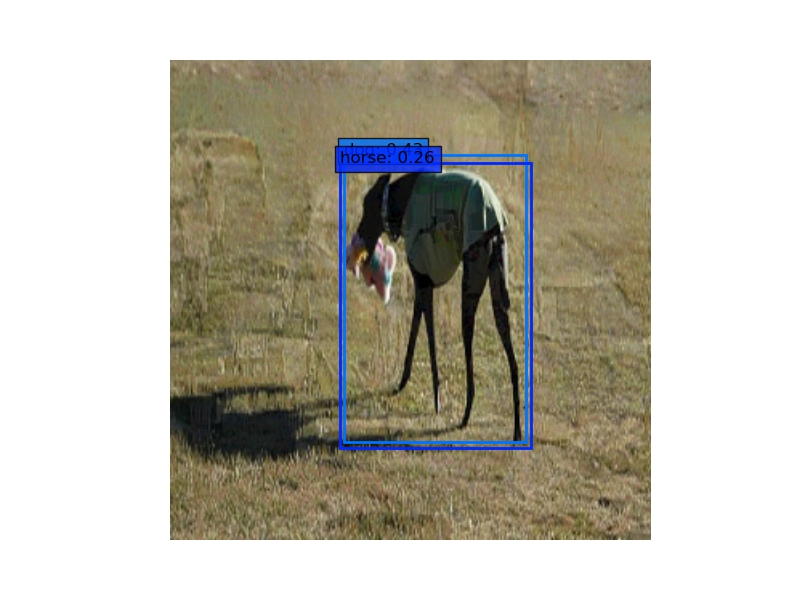}
	\end{overpic}
	\begin{overpic}[viewport=160 60 640 555, clip, width=2.1cm]{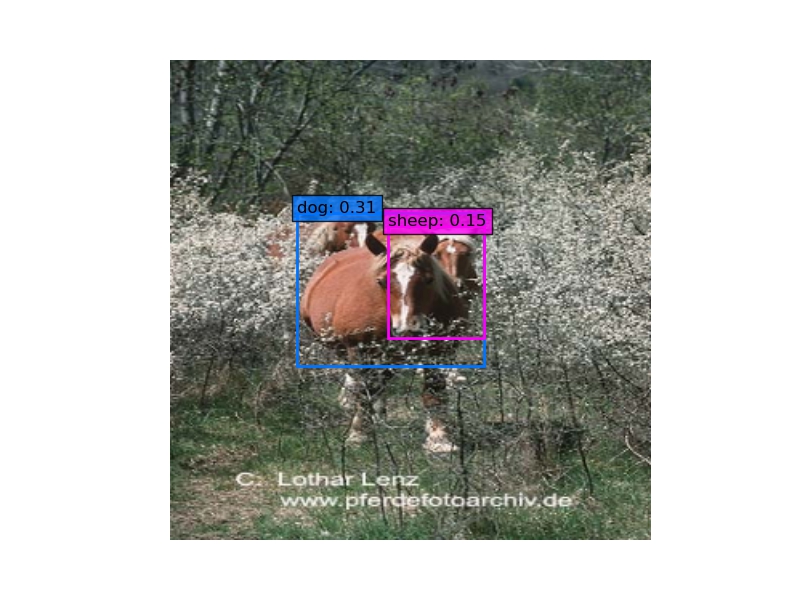}
	\put(90,205){\sffamily\textcolor{black}{{\scalebox{0.75}{\rotatebox{0}{incorrect bounding box and/or class}}}}}
	\end{overpic}
	\begin{overpic}[viewport=160 60 640 555, clip, width=2.1cm]{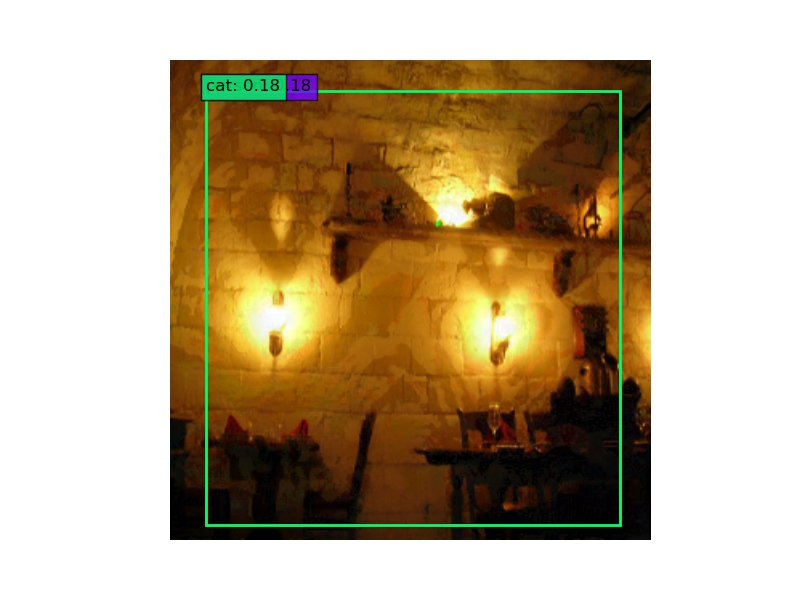}
	\end{overpic}	
	\begin{overpic}[viewport=160 60 640 555, clip, width=2.1cm]{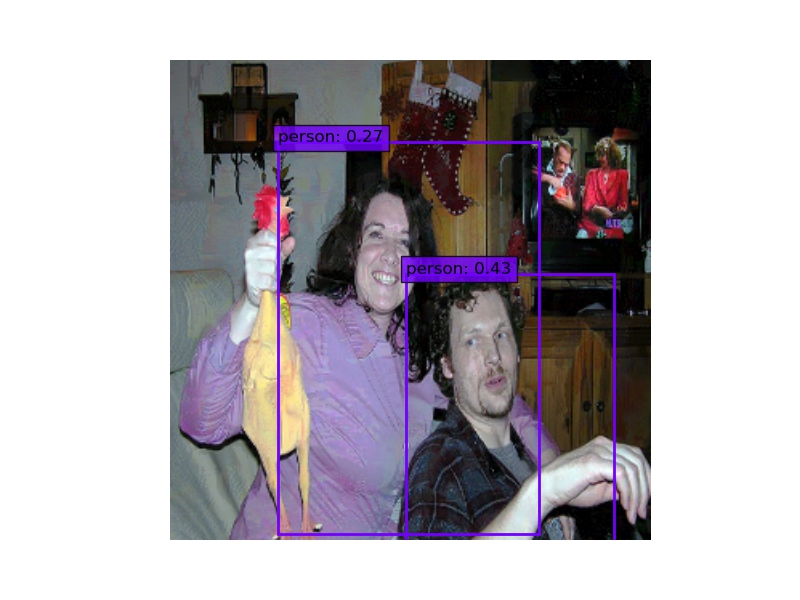}
	\end{overpic}
	\caption{\textbf{Visualization of failure cases}. Example challenging cases include images with small objects and visually confusing classes.}
	\label{fig:failure_case}
	\vspace{-0.18in}
\end{figure*}

\subsection{Defense against Transferred Attacks}
We further test  the performance of the robust models under transferred attacks: attacks that are transferred from models with different 
backbones and/or detection architectures.
Our model under test is based on SSD+VGG16.
For  attacks transferred from different backbones, they are generated under the SSD architecture  but replacing the VGG backbone with ResNet or DarkNet.
For attacks transferred from different detection architectures, we use RFB~\cite{RFB}, FSSD~\cite{FSSD} and YOLO~\cite{yolo, YOLOv3}.\footnote{As the input image size for YOLO is $416\!\times \! 416$, which is different from input size of $300\!\times \! 300$ for SSD, we insert a differentiable interpolation module ($300^2\!\!\rightarrow \!\!416^2$) between the input with size of  $300\!\times \! 300$  and YOLO.}
\alg{DAG}~\cite{DAG} and \alg{RAP}~\cite{BMVC_RAP} are used as the underlining attack generation algorithms.
The results are summarized in Table~\ref{tab:blackbox}.
It is observed that the proposed model is robust against transferred attacks generated with different algorithms and architectures.
It is also observed that the attacks have a certain level of robustness can be transferred across  detectors with different  backbones or structures.
This reconfirms the results from~\cite{DAG, BMVC_RAP}.

\begin{table}[t]
	\centering
	\begin{tabular}{M{2.5cm}|M{1.2cm}|M{1.2cm}|M{0.8cm}}
		{transferred attack} & \salg{DAG}~\cite{DAG} & \centering \salg{RAP}~\cite{BMVC_RAP} &  average  \\
		\Xhline{2\arrayrulewidth}
		{\footnotesize  SSD+ResNet50}  &  49.3   & 49.4 &  49.4 \\
		{\footnotesize SSD+DarkNet53}  &  49.2  &  49.4 & 49.3\\
		{\footnotesize  RFB+ResNet50} &   49.1 &  49.3 &  49.2 \\
		{\footnotesize  FSSD+DarkNet53} & 49.3  &  49.2 & 49.3 \\
		{\footnotesize YOLO+DarkNet53}  & 49.5 &  49.5 &  49.5\\
		%\Xhline{1\arrayrulewidth}
		%average attack & & &
	\end{tabular}
	\caption{Performance of our model (SSD+VGG16) against  attacks transferred from different backbones and detector architectures.}
	\label{tab:blackbox}
	\vspace{-0.15in}
\end{table}

\subsection{Results on MS-COCO}

We further conduct experiments on MS-COCO~\cite{COCO}, which is more challenging both for the standard detector as well as the defense due to its increased number of classes and data variations.
The results of different models under RAP attack~\cite{BMVC_RAP} with attack budget 8 and PGD step 20 are summarized in Table~\ref{tab:task_coco}. 
The standard model achieves a very low accuracy  in the presence of attack (compared with $\mathtt{\sim}$40\% on clean images).
Our proposed models improves over the standard model significantly and performs generally well across different backbones and detection architectures.
This further demonstrates the effectiveness of the proposed approach on improving model robustness.

\subsection{Failure Case Analysis}
We visualize in Figure~\ref{fig:failure_case} some example cases that are challenging to our current model.
Images with small objects that are challenging for the standard detectors~\cite{ssd,yolo} remain to be one category of challenging  examples for robust detectors. Better detector architectures might be necessary to address this challenge. 
Another challenging category is objects with visually confusing appearance, which naturally leads to low confidence predictions.
This is more related to the classification task of the detector and can   benefit from advances in  classification~\cite{feature_dn}.
There are also cases where the predictions are inaccurate or completely wrong, which reveals the remaining challenges in robust detector training.

\begin{table}[t]
	\centering
	\begin{tabular}{M{1.5cm} |M{0.8cm}  M{1.4cm} V{1} R{0.8cm}  R{0.8cm}}
		\centering model & {\small architec.} & {\small backbone} &  clean & attack   \\
		\Xhline{2\arrayrulewidth}
		\salg{standard} &   {\small SSD} & {\small VGG16} & 39.8 & 2.8 \\
		\Xhline{1\arrayrulewidth}
		\multirow{6}{*}{\salg{ours}}   &  {\small SSD} & {\small VGG16} & 27.8&16.5  \\
		& {\small SSD} & {\small DarkNet53} & 20.9  &18.8 \\
		&  {\small SSD} & {\small ResNet50} &  18.0 & 16.4 \\
		&  {\small RFB} & {\small ResNet50} &  24.7  &21.6 \\
		&  {\small FSSD} & {\small DarkNet53} & 23.5   & 20.9\\
		&  {\small YOLO} & {\small DarkNet53} & 24.0 & 21.5  \\
	\end{tabular}
	\caption{Comparison of standard and robust models on MS-COCO under RAP attack~\cite{BMVC_RAP} with attack budget 8 and 20 PGD steps. }
	\label{tab:task_coco}
	\vspace{-0.2in}
\end{table}

\section{Conclusions}
We have presented an approach for improving the robustness object detectors against adversarial attacks. 
From a  multi-task view of object detection, we systematically analyzed  existing attacks for object detectors and the impacts of individual task component on model robustness. An adversarial training method for robust object detection is developed based on these analyses.
Extensive experiments have been conducted on  PASCAL-VOC and MS-COCO datasets and experimental results have demonstrated the efficacy of the proposed approach on improving model robustness compared with the standard model, across different attacks, datasets, detector backbones and architectures.

This work  serves as an initial step towards  adversarially robust detector training with promising results.
More efforts need to be devoted in this direction to address the remaining challenges.
New advances on object detection can be used to further improve the model performance, \emph{e.g.}, better loss function for approximating the true objective~\cite{focal_loss} and different architectures for addressing small object issues~\cite{Cui2018MDSSDMD, DSSD}.
Similarly, as a component task of object detection, any advances on  classification task could be potentially transferred as well~\cite{feature_dn}.
There is also a trade-off between accuracy on clean image and robustness for object detection as in the classification case~\cite{tsipras2018robustness}. 
How to leverage this trade-off better is another future work. 
Furthermore, by viewing object detection as an instance of multi-task learning task, this work could serve as an example on robustness improvement for other multi-task learning problems as well~\cite{kendall2017multi, Zhao_2018_ECCV}.

{\small
\bibliographystyle{ieee}
\bibliography{ARD}
}

\end{document}